\documentclass{article}

\PassOptionsToPackage{numbers, compress}{natbib}

\usepackage[preprint]{neurips_2025}

\usepackage[utf8]{inputenc} 
\usepackage[T1]{fontenc}    
\usepackage{hyperref}       
\usepackage{url}            
\usepackage{booktabs}       
\usepackage{multirow}
\usepackage{tabularx}
\usepackage{wrapfig}
\usepackage{amsfonts}       
\usepackage{nicefrac}       
\usepackage{microtype}      
\usepackage{xcolor}         
\usepackage{graphicx}
\usepackage{epstopdf}
\usepackage{subcaption}
\usepackage[normalem]{ulem}
\useunder{\uline}{\ul}{}

\usepackage{listings}
\usepackage{xcolor}
\lstdefinestyle{cypher}{
  basicstyle=\ttfamily\small,
  backgroundcolor=\color{gray!10},
  frame=single,
  keywordstyle=\color{blue},
  commentstyle=\color{gray},
  stringstyle=\color{orange},
  showstringspaces=false,
  breaklines=true,
}

\usepackage{natbib}
\usepackage{amsmath}
\usepackage{amssymb}
\usepackage{mathtools}
\usepackage{amsthm}

\newtheorem{proposition}{Proposition}

\usepackage{pifont}

\usepackage[capitalize,noabbrev]{cleveref}

\usepackage[textsize=tiny]{todonotes}

\newcommand{\name}{{PolyG}} 
\newcommand{\benchmarkname}{{PolyBench}} 

\newcommand{\stitle}[1]{\vspace*{0.4em}\noindent{\bf #1\/}}
\newcommand{\squishlist}{
  \begin{list}{$\bullet$}
    { \setlength{\itemsep}{1pt}
      \setlength{\parsep}{1pt}
      \setlength{\topsep}{2.5pt}
      \setlength{\partopsep}{0.5pt}
      \setlength{\leftmargin}{1em}
      \setlength{\labelwidth}{1em}
      \setlength{\labelsep}{0.6em}
    }
  }
  \newcommand{\squishend}{
  \end{list}
}

\newcommand*\circled[1]{\tikz[baseline=(char.base)]{
            \node[shape=circle,fill,inner sep=1pt] (char) {\textcolor{white}{#1}};}}

\title{PolyG: Adaptive Graph Traversal for Diverse GraphRAG Questions}

\author{
Renjie Liu$^1$, Haitian Jiang$^2$, Xiao Yan$^3$, Bo Tang$^1$, Jinyang Li$^2$ \\
Southern University of Science and Technology$^1$, New York University$^2$, Wuhan University$^3$ \\
\texttt{liurj2023@mail.sustech.edu.cn}
}

\begin{document}

\maketitle

\begin{abstract}
GraphRAG enhances large language models (LLMs) to generate quality answers for user questions by retrieving related facts from external knowledge graphs. However, current GraphRAG methods are primarily evaluated on and overly tailored for knowledge graph question answering (KGQA) benchmarks, which are biased towards a few specific question patterns and do not reflect the diversity of real-world questions. To better evaluate GraphRAG methods, we propose a \textit{complete four-class taxonomy} to categorize the basic patterns of knowledge graph questions and use it to create \benchmarkname{}, a new GraphRAG benchmark encompassing a \textit{comprehensive} set of graph questions. With the new benchmark, we find that existing GraphRAG methods fall short in \textit{effectiveness} (i.e., quality of the generated answers) and/or \textit{efficiency} (i.e., response time or token usage) because they adopt either a \textit{fixed} graph traversal strategy or \textit{free-form} exploration by LLMs for fact retrieval. However, different question patterns require distinct graph traversal strategies and context formation. To facilitate better retrieval, we propose \name{}, an \textit{adaptive} GraphRAG approach by decomposing and categorizing the questions according to our proposed question taxonomy. Built on top of a unified interface and execution engine, \name{} \textit{dynamically} prompts an LLM to generate a graph database query to retrieve the context for each decomposed basic question. Compared with SOTA GraphRAG methods, \name{} achieves a higher win rate in generation quality and has a low response latency and token cost. Our code and benchmark are open-source at \url{https://github.com/Liu-rj/PolyG}.
\end{abstract}

\section{Introduction}\label{sec:intro}
Large Language Models (LLMs)~\cite{llama2,openai2024gpt4technicalreport,brown2020languagemodelsfewshotlearners} have achieved remarkable success for various natural language processing (NLP) tasks~\cite{multitask, chatbot, translation}. However, they are susceptible to hallucinations when generating answers for questions that require information beyond their knowledge~\cite{hallucination, zhang2023sirenssongaiocean}. To tackle this problem, graph-based retrieval-augmented generation (GraphRAG)~\cite{ms_graphrag} has emerged as an effective approach, which grounds LLM's responses using external knowledge graphs~\cite{ecommerce, finance, social}. When answering a user question, GraphRAG first retrieves the relevant entities and relations from the knowledge graph and then utilizes this information along with the question to prompt the LLM to generate the answer.


\stitle{Limitation of existing GraphRAG evaluation.} While many GraphRAG solutions are proposed, they are primarily evaluated on Knowledge-Graph Question-Answering (KGQA) benchmarks. The questions in these benchmarks mainly ask about the concrete attributes of target entities with specific relation constraints on the related entities. As a result, these benchmarks cannot capture the diversity of real-world graph questions as shown in \autoref{tab:benchmark_comparison}, based on our question taxonomy in~\autoref{sec:question_categorization}. 
The lack of question diversity leads new GraphRAG methods to overfit  specific question patterns in their development and thus hinders their ability to excel in real-world applications. 

To enable more comprehensive evaluation, we introduce a new GraphRAG benchmark called \emph{\benchmarkname{}} (shown in ~\autoref{tab:benchmark_comparison}) that includes a \textit{comprehensive} and \textit{complete} set of question patterns. To do so, we create a taxonomy of graph question patterns by abstracting graph question using a fact triple $\langle$\textit{subject}, \textit{predicate chain}, \textit{object}$\rangle$ and classifying questions \textit{depending on the missing component} in the triple. Under this taxonomy, a question can inquire about specific or unspecified attributes of an entity, explore potential relations between two entities, or verify the existence of a relation. Moreover, our benchmark also includes nested questions consisting of multiple basic questions from the taxonomy to assess the ability of GraphRAG methods on handling complex questions.

\stitle{Limitations of existing GraphRAG methods.} Most existing GraphRAG methods either use a \textit{fixed} graph traversal strategy or rely on LLMs for \textit{free-form} exploration. For example, Microsoft GraphRAG~\cite{ms_graphrag} (hereafter referred to as MS\_GraphRAG) and LightRAG~\cite{lightrag} adopt one-hop Breadth-First Search (BFS) to retrieve information from the neighborhood of the extracted entities; RoG~\cite{rog} deduces concrete and faithful reasoning paths starting from the question's entities and use them for traversal; Fast-GraphRAG~\cite{fast_graphrag} and HippoRAG~\cite{hipporag} use the personalized PageRank (PPR) score~\cite{ppr} to select the top paths; while ToG~\cite{tog} and Graph-CoT~\cite{graphcot} invoke the LLM at each traversal step to select the next entity to visit in the knowledge graph. Recently, methods such as G-retriever~\cite{gretriever}, GNN-RAG~\cite{gnnrag} and SubgraphRAG~\cite{subgraphrag} propose trainable retriever that flexibly adapts to each query. However, they need to train the retriever on the KGQA benchmark and thus are not applicable to general GraphRAG scenarios where questions typically do not have a golden ground truth.

Since the traversal strategy of most existing GraphRAG methods is either \textit{fixed} or \textit{free-form} while the question patterns in GraphRAG workloads can be diverse, they suffer from two limitations.


\squishlist
\item \textit{Limited effectiveness}: Using a fixed graph traversal strategy works well for a specific question pattern but falls short for other patterns. For instance, using one-hop neighbor expansion, MS\_GraphRAG and LightRAG suit questions that ask about general aspects of entities but cannot handle questions involving entities that are more than one-hop apart. Relying on the question to deduce the reasoning paths, RoG works well when the question provides explicit relation constraints but fails when these constraints are missing or vague. Fast-GraphRAG and HippoRAG are good for questions without constraints by ranking possible paths using their PPR scores but the selected top-ranking paths may be irrelevant when the question has explicit constraints.

\item \textit{High cost}: By using an LLM to determine the next entity to explore at every traversal step, ToG and Graph-CoT can handle all question patterns. However, they incur significant overhead due to frequent LLM invocations, increasing the end-to-end response time by approximately 2× and the number of used tokens by 10× according to our profiling. Moreover, they often do not produce the best response quality by focusing on fine-grained bootstrapping with a narrow view of knowledge.
\squishend

\begin{table}[!t]
\centering
\caption{Question coverage and data source of our \benchmarkname{} and existing KGQA benchmarks. In the question pattern, $s$ means subject, $p$ is predicate chain, $o$ refers to object, and $*$ is the missing component. WIKIQA inspires our benchmark and targets text-understanding task rather than KGQA.}
\label{tab:benchmark_comparison}
\resizebox{0.75\textwidth}{!}{%
\begin{tabular}{@{}cccccc@{}}
\toprule
\multirow{3}{*}{\textbf{Benchmark}} & \multicolumn{4}{c}{\textbf{Basic Question Pattern}} & \multirow{3}{*}{\textbf{Data Source}} \\ \cmidrule(lr){2-5}
 & \multirow{2}{*}{$\langle s,*,* \rangle$} & \multirow{2}{*}{$\langle s,p,* \rangle$} & \multirow{2}{*}{$\langle s,*,o \rangle$} & \multirow{2}{*}{$\langle s,p,o \rangle$} &  \\
 &  &  &  &  &  \\ \midrule
SimpleQ~\cite{simpleq} (2015) &  & \ding{51} &  &  & Freebase Fact \\
WebQSP~\cite{webqsp} (2016) & \ding{51} & \ding{51} &  &  & Google Suggest API \\
GraphQ~\cite{graphq} (2016) &  & \ding{51} &  &  & Random Construct. \\
CWQ~\cite{cwq} (2018) &  & \ding{51} &  &  & WebSQP \\
GrailQA~\cite{grailqa} (2021) &  & \ding{51} &  &  & Random Construct. \\
RGBench~\cite{graphcot} (2024) &  & \ding{51} &  &  & Question Template \\
CypherBench~\cite{cypherbench} (2025) &  & \ding{51} &  &  & Question Template \\
WIKIQA~\cite{wikiqa} (2015) & \ding{51} & \ding{51} & \ding{51} &  & Bing Query Log \\
\textbf{PolyBench (Ours)} & \ding{51} & \ding{51} & \ding{51} & \ding{51} & Question Template \\ \bottomrule
\end{tabular}%
}
\vspace{-4mm}
\end{table}

\begin{figure*}[!t]
	\centering
\includegraphics[width=\textwidth]{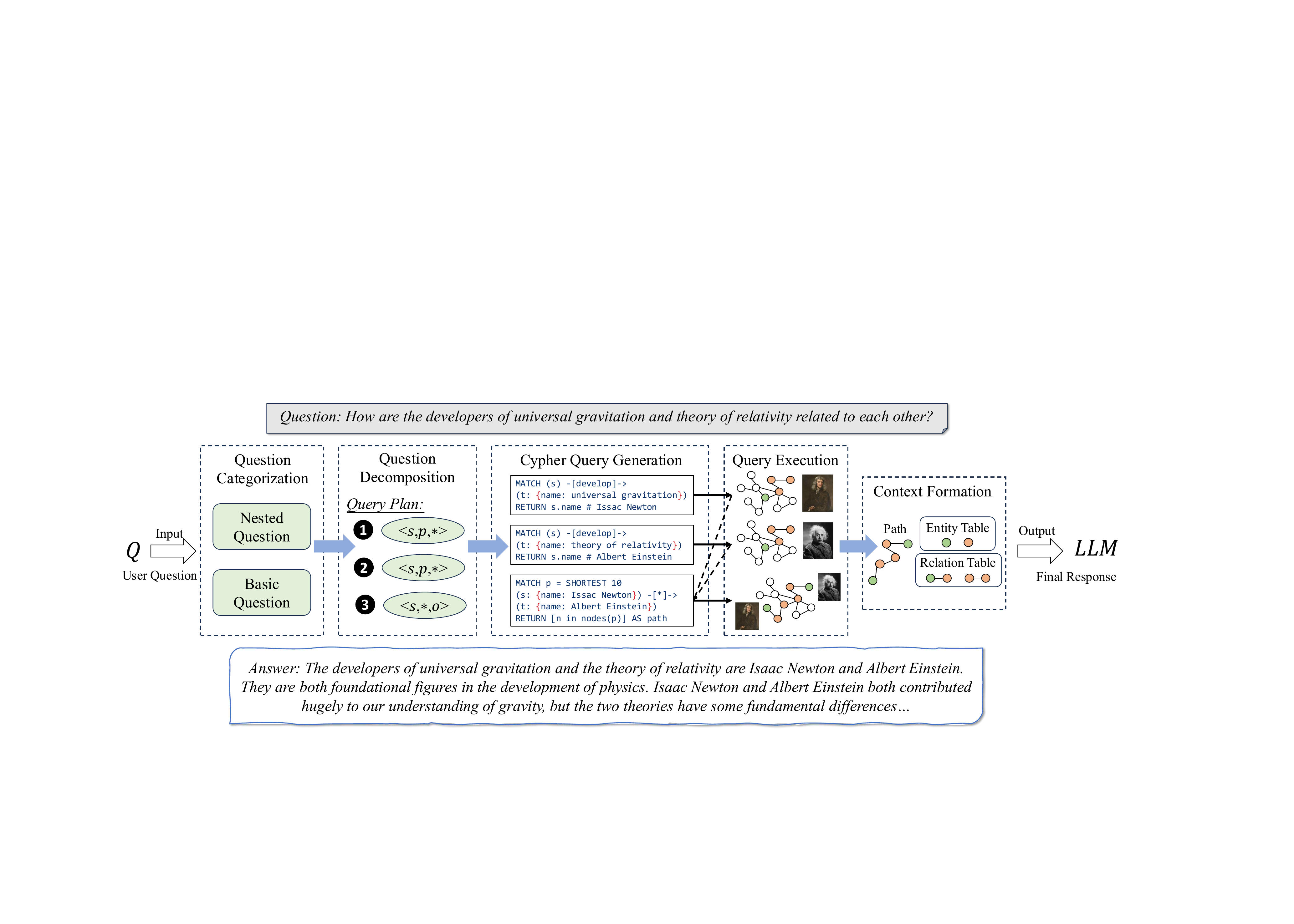}
	\caption{The workflow of our \name{}. \name{} first categorizes the user question according to its complexity (i.e., basic or nested). Then, a query plan is produced where each step is a basic graph question. For each step, \name{} generates a concrete Cypher query to retrieve information from the knowledge graph. Finally, \name{} forms the final context and prompt the LLM to generate  response.}
	\label{fig:workflow}
	 \vspace{-4mm}
\end{figure*}

\stitle{Our solution \name{}.} 
To tackle the limitations of existing solutions, we propose \name{}, which achieves high accuracy for diverse graph questions while maintaining efficiency. As shown in \autoref{fig:workflow}, \name{} first categorizes the input user question into basic or nested questions based on our four-class question taxonomy. Then, for nested questions, \name{} further decomposes the overall question into several basic questions and adaptively prompts the LLM to generate graph database queries to retrieve the relevant information.  In particular, \name{} generates Cypher queries~\cite{cypher} which are supported by many popular graph databases and the retrieved results are organized into a concise and clear format for the LLM to generate a high-quality response. To our knowledge, \name{} is the first query planner for GraphRAG, and by introducing an explicit query planning stage, \name{} also opens the door for 
other advanced query processing strategies, such as query rewriting and merging.


Evaluated on \benchmarkname{}, we compare the GraphRAG methods in terms of answer quality, end-to-end latency, and token usage. Extensive experimental results show that our \name{} consistently outperforms all baselines, achieving the highest win rates in all evaluation criteria for answer quality, and meanwhile reduces the response latency by up to 2× and the token consumption by up to 90\%.

To summarize, we make the following key contributions.

\squishlist
\item We systematically study the query patterns in real graph questions and observe the lack of question diversity in current KGQA benchmarks that are used to evaluate GraphRAG methods.
\item We introduce a new benchmark, \benchmarkname{}, tailored for GraphRAG workloads that encompass a \textit{comprehensive} and \textit{complete} set of question patterns guided by a four-class taxonomy.
\item We develop a general GraphRAG solution, \name{}, that automatically categorizes and decomposes questions into basic queries, and adaptively prompts the LLM to generate the executable Cypher query to retrieve necessary context needed to answer the question.
\squishend

\section{GraphRAG Basics}\label{sec:background}




\stitle{Knowledge graph.}
A knowledge graph is typically defined as $\mathcal{KG} = (\mathcal{V}, \mathcal{E})$, where $\mathcal{V}$ is the set of entities and $\mathcal{E}$ is the set of relations. Each entity $v \in \mathcal{V}$ and relation $e \in \mathcal{E}$ is often associated with specific attributes. For example, in an academic knowledge graph, entities may have types such as ``\textit{paper}'', while relations may represent semantics such as ``\textit{cite}''. Moreover, knowledge graphs commonly include inverse edges to capture mutual relationships. Specifically, for an edge $e = (u, r, v)$, where $u$ and $v$ are entities and $r$ is a relation type, the inverse edge is defined as $e' = (v, r^{-1}, u)$, where $r^{-1}$ denotes the inverse of relation $r$.

\stitle{Reasoning path.} A reasoning path in a knowledge graph is a sequence of entities connected by relations. It is represented as $P = v_0 \rightarrow (e_0) \rightarrow v_1 \rightarrow (e_1) \rightarrow \cdots \rightarrow (e_n) \rightarrow v_n$, where $v_i \in \mathcal{V}$ and $e_i \in \mathcal{E}$ denotes the $i$-th entity and relation. Specifically, each reasoning path corresponds to a fact triple $\langle s,p,o \rangle$, where $s$ is the subject (entity), $p$ is the predicate chain (chain of relations) and $o$ is the object (entity). For example, the reasoning path for the fact ``\textit{Universal gravitation was developed by Issac Newton who was born in 1643.}'' is: \texttt{Universal Gravitation $\rightarrow$ (developer) $\rightarrow$ Isaac Newton $\rightarrow$ (born in) $\rightarrow$ 1643}, where the subject $s$ is ``\textit{Universal Gravitation}'', the predicate chain $p$ is ``\textit{was developed by Issac Newton who was born in}'' and the object $o$ is ``\textit{1643}''.

\stitle{GraphRAG.} Given a natural language question $Q$ and a knowledge graph $KG$, the main challenge facing a GraphRAG solution is to develop a retriever $\psi$ that can extract relevant entities, relations, and reasoning paths from the knowledge graph. The retrieved context is then provided to the LLM to generate the final answer $A$. This process is formalized as:

\begin{equation}\label{equ:graphrag}
A = \mathcal{LLM}(Q, \psi(KG, Q)).
\end{equation}

Existing GraphRAG methods mainly differ in how their retrievers traverse the knowledge graph and extract the relevant facts. Detailed descriptions of their workflow are provided in \autoref{sec:existing_work_appendix}.

\section{Graph Question Patterns}\label{sec:question_categorization}

As is discussed in~\autoref{sec:intro}, the lack of question diversity in existing KGQA benchmarks motivates us to explore more question patterns that can appear in real-world scenarios.

\begin{table*}[!t]
\centering
\caption{The 4 basic and some common nested question patterns in our taxonomy and their descriptions along with concrete examples. In a knowledge graph fact triple $\langle s,p,o \rangle$, $s$ stands for the subject, $p$ refers to the predicate chain, and $o$ is the object.}
\label{tab:question_description}
\resizebox{\textwidth}{!}{%
\begin{tabular}{@{}llm{15cm}@{}}
\toprule
\multicolumn{2}{l}{\textbf{Question Pattern}} & \textbf{Description \& Example} \\ \midrule
\multirow{4}{*}{\raisebox{-5em}{\rotatebox[origin=c]{90}{Basic Pattern}}} & $\langle s,*,* \rangle$ & Questions about an entity (the subject) with no specific relation constraints (the predicate) and target entity (the object). The task is to answer a general question about the entity. Example: \textit{``Who is Isaac Newton?''} \\ \cmidrule(l){2-3}
& $\langle s,p,* \rangle$ & Questions about an entity (the subject) with specific relation constraints (the predicate) but misses the target entity (the object). The task is to answer one specific aspect of the entity. Example: \textit{``What theories and principles has Isaac Newton developed?''} \\ \cmidrule(l){2-3}
& $\langle s,*,o \rangle$ & Questions about any relations (the predicate) between two entities (the subject and object). The task is to provide the relations between  two entities. Example: \textit{``How is Isaac Newton and Albert Einstein related?''} \\ \cmidrule(l){2-3}
& $\langle s,p,o \rangle$ & Questions about specific relations (the predicate) between two entities (the subject and object). The task is to check the existence of a specific relationship between the two entities. Example: \textit{``Have Isaac Newton and Albert Einstein both contributed to the same same field of science?''} \\ \midrule \midrule
\multirow{4}{*}{\raisebox{-4em}{\rotatebox[origin=c]{90}{Nested Pattern}}} & $\langle s,*,* \rangle + \langle s,p,* \rangle$ & Nested questions about general information of an unknown entity with specific relation constraints. Example: \textit{``Tell me about the scientist who developed univsersal gravitation.''} \\ \cmidrule(l){2-3}
& $\langle s,p,* \rangle + \langle s,p,* \rangle$ & Nested questions about an entity with specific but convoluted relation constraints (non-linear chains of relations). Example: \textit{``Who is the developer of universal gravitation and influenced by Galileo's discoveries?''} \\ \cmidrule(l){2-3}
& $\langle s,*,o \rangle + \langle s,p,* \rangle$ & Nested questions about any relations between unknown entities with specific constraints. Example: \textit{``How are the developers of universal gravitation and theory of relativity related to each other?''} \\ \cmidrule(l){2-3}
& $\langle s,p,o \rangle + \langle s,p,* \rangle$ & Nested questions about specific relations between entities with specific constraints. Example: \textit{``Do the developers of universal gravitation and theory of relativity share similar perspectives about gravitation?''} \\ \bottomrule
\end{tabular}%
}
\end{table*}

\stitle{Questions from real queries.} Due to privacy concerns~\cite{aolquerylog}, real-world query logs are rarely made publicly available. However, we studied the WIKIQA benchmark~\cite{wikiqa}, which is sampled from raw Bing query logs and serves as a proxy for real user questions. While existing KGQA benchmarks primarily cover structured factual questions, our analysis of WIKIQA reveals a broader range of question patterns. In addition to factual queries about entities or relationships, users also ask more descriptive or explanatory questions. For example, questions such as ``\textit{What is feedback mechanism in plants during respiration?}'' and ``\textit{How are the directions of the velocity and force vectors related in a circular motion?}'' highlight this diversity. These observations motivate us to develop a \textit{complete} set of question patterns derived from the knowledge graph reasoning paths.

\stitle{Basic question patterns.} Based on the analysis of real graph queries, we observe that questions in knowledge graphs correspond to queries over different components of reasoning paths, aka fact triples $\langle s, p, o \rangle$, where $s$ is the subject, $p$ is the predicate chain, and $o$ is the object. By masking each element of the triple as either known or unknown ($*$), a total of eight possible question patterns ($2^3$) can be generated. For example, a question of the form $\langle s, *, * \rangle$ could be ``\textit{Tell me about Isaac Newton?}''. However, some combinations do not yield meaningful questions, while others are redundant. Specifically, the pattern $\langle *, *, * \rangle$ does not correspond to any valid query. Additionally, the patterns $\langle *, p, o \rangle$ and $\langle *, *, o \rangle$ are equivalent to $\langle s, p, * \rangle$ and $\langle s, *, * \rangle$, respectively, due to the existence of inverse relations in knowledge graphs. The pattern $\langle *, p, * \rangle$ necessitates a full scan of all relations in $\mathcal{KG}$, which typically does not correspond to meaningful questions, and thus is excluded from this work. Consequently, we identify four valid basic question patterns: $\langle s, *, * \rangle$, $\langle s, p, * \rangle$, $\langle s, *, o \rangle$, $\langle s, p, o \rangle$. In \autoref{tab:question_description}, we define each of these four patterns, specify their associated tasks, and provide illustrative examples. These four patterns lead us to the following proposition:

\begin{proposition}
\label{pro:taxonomy}
Any valid graph question either belongs to one of the four basic question patterns 
$\langle s, *, * \rangle$, $\langle s, p, * \rangle$, $\langle s, *, o \rangle$, or 
$\langle s, p, o \rangle$, or can be decomposed into a set of sub-questions, each of 
which falls into one of these four basic patterns.
\end{proposition}

Based on the taxonomy, we conducted a comprehensive analysis of the question patterns covered by existing KGQA benchmarks, as summarized in \autoref{tab:benchmark_comparison}. Despite differences in their data curation processes, we observe that these benchmarks exhibit limited diversity in question types, typically covering only 2 out of the 8 patterns. Further details are provided in \autoref{sec:kbqa_benchmark_appendix}. This lack of diversity limits the ability to evaluate GraphRAG’s general capability in addressing real-world questions. It is worth noting that, WIKIQA does not cover all of the proposed patterns, because it samples a limited subset of queries from Bing logs that begin with a WH-word (e.g., ``what'' or ``how''), and therefore does not fully reflect the diversity or distribution of questions in real-world scenarios.

\begin{figure*}[!t]
	\centering
\includegraphics[width=\textwidth]{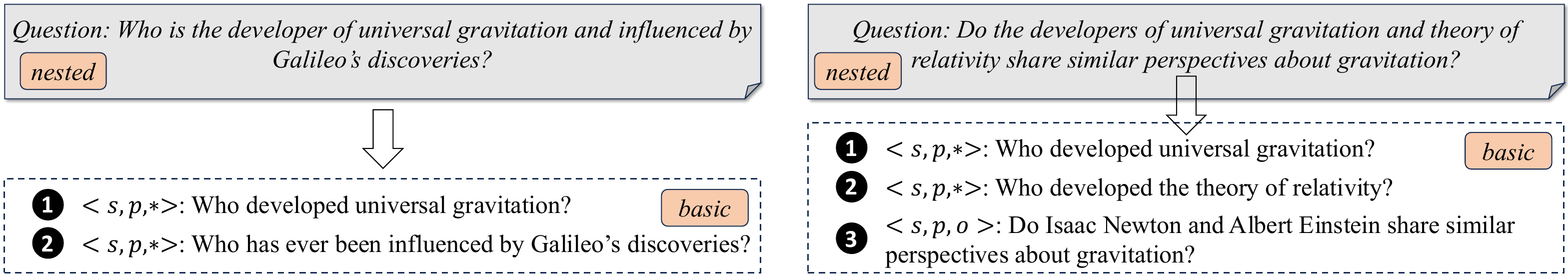}
	\caption{Examples of how nested questions can be decomposed into several basic questions.}
\label{fig:nested_question}
	 \vspace{-4mm}
\end{figure*}

\stitle{Nested questions.} In real-world scenarios, natural language questions over graphs can be complex and involve the nesting of multiple basic questions. For instance, recent KGQA benchmarks such as CWQ~\cite{cwq}, GraphQ~\cite{cwq}, and GrailQA~\cite{grailqa} include questions with intricate fact constraints that go beyond a simple linear chain of relations. Consider the question: ``\textit{Who is the developer of both universal gravitation and the law of motion?}'' This can be decomposed into two basic $\langle s, p, * \rangle$ sub-questions: ``\textit{Who developed universal gravitation?}'' and ``\textit{Who formulated the law of motion?}''. We observe that only questions with the $\langle s, p, * \rangle$ pattern can serve as internal sub-questions in nested queries, as they yield specific answers that can be passed to the outer question. Building on this, we identify three common nested question patterns: $\langle s, *, * \rangle + \langle s, p, * \rangle$, $\langle s, *, o \rangle + \langle s, p, * \rangle$, and $\langle s, p, o \rangle + \langle s, p, * \rangle$, which are composed of multiple basic question patterns. Definitions and examples of these nested question patterns are provided in \autoref{tab:question_description}. Furthermore, \autoref{fig:nested_question} illustrates how nested questions can be decomposed into basic sub-questions.

Note that our taxonomy still may not capture all possible knowledge graph questions, particularly given the high flexibility of natural language. For example, questions can require global summaries~\cite{ms_graphrag} or combine more than two basic question patterns. However, such cases are relatively rare and still conform to Proposition~\ref{pro:taxonomy}—they can be decomposed into sub-questions that follow our four basic question patterns.

\section{\benchmarkname: A Benchmark for GraphRAG with Diverse Questions}\label{sec:benchmark}

In this section, we first introduce the knowledge graphs used in \benchmarkname, then discuss how we generate concrete graph questions and paraphrase them into flexible expressions.


\stitle{Data collection.} We utilize knowledge graphs from the public graph reasoning benchmark GRBench~\cite{graphcot}, which contains 10 knowledge graphs spanning 5 general domains: academia, e-commerce, literature, healthcare, and legal, collected from various public sources. In \benchmarkname{}, we select 3 out of the 10 graphs—specifically from the academia~\cite{academia}, literature~\cite{literature}, and e-commerce~\cite{e-commerce} domains. Detailed statistics of the selected knowledge graphs are provided in \autoref{sec:dataset_appendix}.

\stitle{Question generation.} Following principles used in prior benchmarks~\cite{cwq, graphcot, grailqa, cypherbench}, we generate initial fixed-form questions by randomly selecting entities that meet specific constraints and populating predefined question templates. Specifically, we design 73 well-crafted question templates across all three knowledge graphs, covering both basic and nested question patterns with varying levels of complexity. Each template is paired with a manually written Cypher query to retrieve entities that ensure the resulting questions are valid and answerable. In particular, inspired by the diverse complexity of $\langle s, p, * \rangle$ questions seen in recent KGQA benchmarks~\cite{cwq, graphq, grailqa, graphcot}, our templates include questions requiring multi-hop reasoning, with relation paths ranging from 1 to 5 hops. For higher-order relations, we carefully validate the semantics to avoid generating unrealistic or nonsensical questions. As a result, we generate 400 initial questions per knowledge graph, covering a broad range of patterns and complexities—yielding a total of 1,200 questions across the entire benchmark. The question templates are described in detail in \autoref{sec:question_template_appendix}.


\stitle{Partial answer annotation.} Most questions in \benchmarkname{} do not have gold-standard ground truth answers, and LLM-based judgment is the common practice for evaluation~\cite{ms_graphrag, fast_graphrag}. Nevertheless, we identify that ground truth can be provided for specific question types—namely, $\langle s, p, * \rangle$ and $\langle s, p, * \rangle + \langle s, p, * \rangle$ questions. For these templates, we manually author additional Cypher queries that, when populated with the selected entities during question generation, retrieve the corresponding answers. These ground truth answers enable quantitative evaluation using standard metrics such as \textit{F1-score} and \textit{Hit}, thereby strengthening the rigor of our benchmark assessment.

\stitle{Question paraphrasing.} As noted in existing KGQA benchmarks~\cite{cwq, graphq, grailqa, graphcot, cypherbench}, template-based question generation often results in fixed and repetitive phrasing, which limits linguistic diversity. To address this, we follow the strategy adopted by recent benchmarks~\cite{graphcot, cypherbench} and use an LLM (Claude-3.5-Sonnet~\cite{claude}) to paraphrase each initial question into four distinct variants. We then randomly select one of these paraphrased versions as the final question presented in \benchmarkname{}.

\section{The \name{} System}\label{sec:solution}

In this section, we introduce \name{}, a general and effective GraphRAG solution designed to handle a wide range of graph questions. \name{} automatically classifies and decomposes user questions into a sequence of basic sub-questions. For each sub-question, it generates an appropriate Cypher query with adaptive prompting and self-correction mechanisms to retrieve relevant information. The responses to these sub-questions are then aggregated and fed into the LLM to produce a final answer.

As depicted in~\autoref{fig:workflow}, \name{} comprises five main stages: \textit{question categorization}, \textit{question decomposition}, \textit{Cypher query generation}, \textit{query execution} and \textit{context formation}. In the following, we describe each stage in detail. The full prompt used in each module is provided in \autoref{sec:prompts_appendix}.

\stitle{Question categorization.} Given a user question, \name{} prompts the LLM to classify it as either a \textit{basic} or \textit{nested} question by providing clear definitions of the basic question patterns along with few-shot examples. A basic question directly corresponds to one of the four defined patterns, whereas a nested question represents a composition of multiple basic questions. If the input is identified as a basic question, the LLM is also required to specify its exact pattern type. For instance, the question “\textit{Who developed universal gravitation?}” is categorized as a basic question with the pattern $\langle s, p, * \rangle$.

\begin{figure*}[!t]
	\centering
\includegraphics[width=0.9\textwidth]{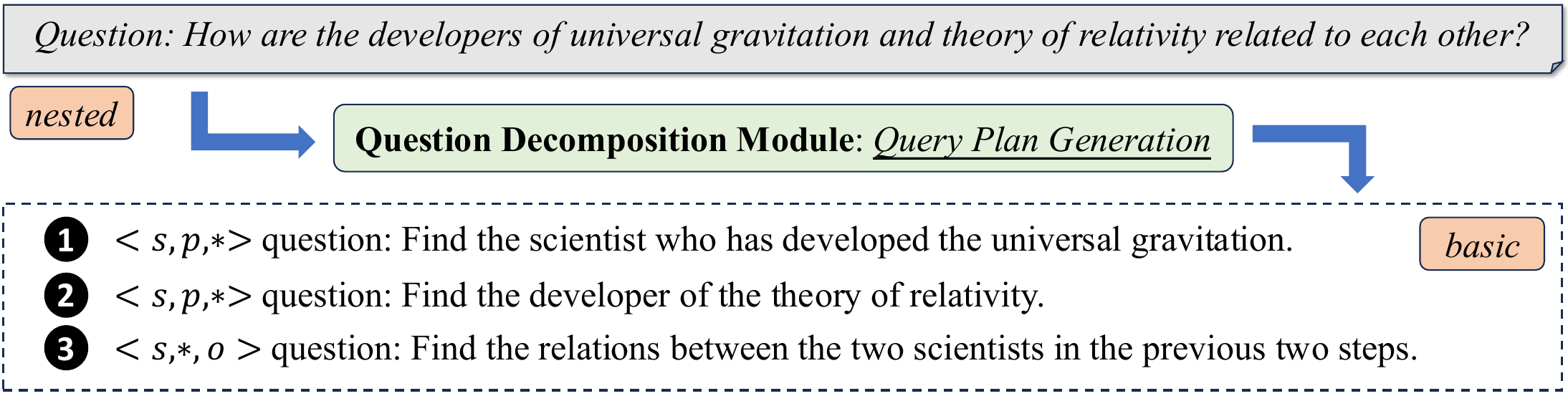}
	\caption{An example of the question decomposition in \name{}.}
\label{fig:question_decompose}
	 \vspace{-4mm}
\end{figure*}

\stitle{Question decomposition.} For nested questions, \name{} prompts the LLM to decompose the complex input into a sequence of basic questions. However, it is often not possible to generate concrete sub-questions without first obtaining intermediate results. For instance, given the question “\textit{How are the developers of universal gravitation and the theory of relativity related to each other?}”, we cannot form the sub-question “\textit{How are Isaac Newton and Albert Einstein related?}” until we have identified these individuals as the developers in earlier steps. Therefore, rather than asking the LLM to generate fully instantiated sub-questions, we prompt it to produce a \textit{query plan}, where each step describes a specific sub-question to be resolved. An example is shown in~\autoref{fig:question_decompose}, where the third step in the query plan is: “\textit{Find the relations between the two scientists identified in the previous two steps},” rather than a fully specified question.

\stitle{Cypher query generation.} Once the query plan is constructed with descriptive steps, \name{} instantiates a concrete basic question for each step before execution. At this stage, the LLM is given the full query plan along with all previously generated sub-questions and their responses, and is asked to generate a concrete question for the current step. \name{} then adaptively prompts the LLM to produce a faithful Cypher query by providing the graph schema (i.e., entity and relation types) and task-specific instructions. While the full schemas of large RDF knowledge graphs (e.g., Wikidata and Freebase) may exceed the LLM's context length, prior work~\cite{cypherbench,rdf2property} has proposed techniques to convert RDF graphs into multiple smaller property graphs with manageable schema sizes. Since each of the four basic question patterns targets different components of a knowledge graph fact triple, as discussed in~\autoref{sec:question_categorization}, we observe that they necessitate \textit{distinct graph traversal strategies}.

\begin{wraptable}{r}{0.45\textwidth}
\centering
\caption{Correspondence from graph question patterns to specific graph traversal strategies.}
\label{tab:traversal_patterns}
\resizebox{0.45\textwidth}{!}{%
\begin{tabular}{@{}cc@{}}
\toprule
\textbf{Question Pattern} & \textbf{Graph Traversal Strategy}         \\ \midrule
$\langle s,*,* \rangle$ & BFS neighbor expansion    \\
$\langle s,p,* \rangle$ & meta-path guided walk      \\
$\langle s,*,o \rangle$ & top-k shortest paths             \\
$\langle s,p,o \rangle$ & top-k constrained shortest paths \\ \bottomrule
\end{tabular}%
}
\end{wraptable}

In~\autoref{tab:traversal_patterns}, we summarize the graph traversal strategies employed for each basic question pattern. Specifically, $\langle s,*,* \rangle$ questions aim to retrieve general information and are best handled via \textit{BFS neighbor expansion}, using queries such as \texttt{``MATCH (s)-[p]->(o) RETURN p, o''}. This approach efficiently captures the immediate neighborhood of the subject $s$ by gathering its one-hop neighbors and directly connected relations. For $\langle s,p,* \rangle$ questions, where the relation $p$ is explicitly defined, we prompt the LLM to generate \textit{concrete meta-paths} that connect the subject $s$ to the target entity $o$. A typical Cypher query for this pattern is of the form \texttt{``MATCH (s)-[p1]->(o1)-[p2]->...->(on) WHERE... RETURN on''}, where \texttt{n} denotes the number of hops in the relational chain and $p_1...p_n$ constitute $p$. This guided traversal not only accurately locates the answer but also reduces unnecessary context. For $\langle s,*,o \rangle$ and $\langle s,p,o \rangle$ questions, which inquire about the relations, we generate Cypher queries that perform a \textit{top-$k$ shortest paths} search to identify all relevant paths between $s$ and $o$, using queries such as \texttt{``MATCH P = SHORTEST k (s)-[*]->(o) RETURN P''}. In the case of $\langle s,p,o \rangle$ questions, where $p$ is known, we further prompt the LLM to apply filtering clauses (e.g., \texttt{``WHERE''}) to ensure that only paths matching $p$ are returned. Beyond shortest-paths, \name{} can also be extended to use random walk methods such as DeepWalk~\cite{deepwalk} and Node2Vec~\cite{node2vec} to uncover meaningful reasoning paths.

\stitle{Query execution.} After the LLM generates a concrete, executable Cypher query for each basic question, \name{} executes the query asynchronously over the underlying graph database. However, even with clear specifications and few-shot examples tailored to each basic question pattern, the LLM may still generate erroneous Cypher queries—such as those with incorrect syntax or nonexistent relations—which can lead to execution errors or empty results. To address this, \name{} incorporates a \textit{self-correction} mechanism that enables the LLM to automatically detect and refine faulty queries upon failure. Specifically, \name{} captures the execution error and sends it back to the LLM along with prior responses to prompt correction. Empirically, we find that setting the maximum number of retries to 3 strikes a good balance between quality and efficiency.

\stitle{Context formation.} Based on the results of each Cypher query from the previous stage, \name{} assembles a complete context to enable the LLM to generate a high-quality response. In addition to task-specific prompts, the core components of this context are the entity and relation tables. The entity table lists all attributes of the retrieved entities (e.g., node type and description), while the relation table aggregates all existing relations and their attributes between each pair of entities. Together, these two tables form the essential context that contains all necessary information to answer the question. Furthermore, for $\langle s, *, o \rangle$ and $\langle s, p, o \rangle$-type questions, we also include additional complete reasoning paths between the question entities to better demonstrate their interconnections. Finally, once \name{} completes all steps in the query plan, it compiles all sub-questions, their corresponding responses, and the original user question to prompt the LLM to generate a final summary.

\section{Evaluation}\label{sec:evaluation}

In this part, we conduct experiments to evaluate \name{} and compare it with existing baseline solutions. 




\begin{table}[!t]
\caption{Response generation quality and efficiency of our \name{} and the baselines on \benchmarkname{}. The win rates sum over 100\% as we allow multiple winners for each question. \textbf{Bold faces} denote the best results. F1-score and Hit are only for $\langle s,p,* \rangle$ and $\langle s,p,* \rangle + \langle s,p,* \rangle$ questions.}
\label{tab:main_results}
\resizebox{\textwidth}{!}{%
\begin{tabular}{@{}llcccccc@{}}
\toprule
\multicolumn{2}{l}{\textbf{Criteria}} & \textbf{MS\_GraphRAG} & \textbf{RoG} & \textbf{Fast-graphrag} & \textbf{Graph-CoT} & \textbf{Cypher} & \textbf{PolyG} \\ \midrule
\multirow{9}{*}{\raisebox{-4em}{\rotatebox[origin=c]{90}{Claude-3.5-sonnet}}} & Comprehensiveness & 26.08\% & 28.75\% & 17.17\% & 19.42\% & 13.25\% & \textbf{74.67\%} \\
 & Diversity & 19.00\% & 26.33\% & 10.67\% & 15.83\% & 8.25\% & \textbf{66.08\%} \\
 & Empowerment & 24.42\% & 26.25\% & 15.83\% & 13.58\% & 12.67\% & \textbf{64.83\%} \\
 & Directness & 29.83\% & 15.75\% & 31.83\% & 42.42\% & 15.50\% & \textbf{46.67\%} \\
 & Overall Winner & 17.17\% & 19.25\% & 12.25\% & 13.83\% & 7.92\% & \textbf{54.75\%} \\ \cmidrule(l){2-8} 
 & F1-score & 0.2094 & 0.6934 & 0.3304 & 0.3885 & 0.1750 & \textbf{0.7084} \\
 & Hit & 0.4400 & 0.8867 & 0.6500 & 0.6500 & 0.3300 & \textbf{0.8900} \\ \cmidrule(l){2-8} 
 & Latency (s) & \textbf{11.79} & 16.75 & 45.84 & 53.81 & 14.84 & 25.38 \\
 & Token Usage & 17,464 & \textbf{1,417} & 48,038 & 104,594 & 4,273 & 6,950 \\ \midrule \midrule
\multirow{9}{*}{\raisebox{-3em}{\rotatebox[origin=c]{90}{Deepseek-R1}}} & Comprehensiveness & 29.17\% & 30.00\% & 33.08\% & 9.17\% & 22.83\% & \textbf{62.83\%} \\
 & Diversity & 24.00\% & 24.00\% & 29.67\% & 4.17\% & 14.92\% & \textbf{56.75\%} \\
 & Empowerment & 29.75\% & 25.50\% & 31.00\% & 4.08\% & 20.42\% & \textbf{50.58\%} \\
 & Directness & 33.00\% & 22.33\% & 36.58\% & 25.17\% & 29.83\% & \textbf{43.75\%} \\
 & Overall Winner & 21.92\% & 21.83\% & 24.17\% & 5.08\% & 17.50\% & \textbf{42.92\%} \\ \cmidrule(l){2-8} 
 & F1-score & 0.3246 & 0.6048 & 0.4752 & 0.2952 & 0.2835 & \textbf{0.6153} \\
 & Hit & 0.4400 & 0.6767 & 0.7000 & 0.3867 & 0.3533 & \textbf{0.6867} \\ \cmidrule(l){2-8} 
 & Latency (s) & \textbf{13.50} & 17.05 & 46.76 & 93.02 & 23.03 & 22.40 \\
 & Token Usage & 16,843 & \textbf{1,534} & 48,038 & 86,128 & 5,014 & 5,796 \\ \bottomrule
\end{tabular}%
}
\end{table}

\subsection{Experiment Setting}

We compare \name{} against several baselines: MS\_GraphRAG~\cite{ms_graphrag}, RoG~\cite{rog}, Fast-graphrag~\cite{fast_graphrag}, and Graph-CoT~\cite{graphcot}, on \benchmarkname{}. Additionally, we include a baseline named \textit{Cypher}, which directly generates Cypher queries using the LLM without guidance from our proposed question pattern taxonomy to show the necessity of our adaptive prompting. To evaluate performance, we follow the evaluation practices of previous work~\cite{ms_graphrag, lightrag}, assessing generation quality based on win rates judged by the LLM across five dimensions: \textit{Comprehensiveness}, \textit{Diversity}, \textit{Empowerment}, \textit{Directness}, and \textit{Overall Winner}. Following previous works, we also report \textit{F1-score} and \textit{Hit} on the questions having golden ground-truth, namely of the type $\langle s,p,* \rangle$ and $\langle s,p,* \rangle + \langle s,p,* \rangle$, along with \textit{response latency} and \textit{token usage} as efficiency metrics. We evaluate the main results using both Claude-3.5-sonnet~\cite{claude} and Deepseek-R1~\cite{deepseekr1} for generation, and Deepseek-R1 to give judgments. For ablation studies and micro-benchmarks, we provide the results with Claude-3.5-sonnet. Detailed experimental settings are provided in~\autoref{sec:exp_settings_appendix}.

\subsection{Experiment Results}

\stitle{Main results.} \autoref{tab:main_results} presents the generation quality and execution efficiency across all three knowledge graphs in \benchmarkname{}. In terms of generation quality, \name{} consistently achieves the highest win rates across all four evaluation criteria, with the most overall wins, as well as the highest F1-score and Hit. These results highlight the limitations of existing GraphRAG methods and underscore the importance of adaptive graph traversal in handling diverse graph questions. The superior performance of \name{} on both Claude-3.5-sonnet and Deepseek-R1 suggests its generalization across different LLMs. RoG achieves comparable quantitative scores to \name{}, as it is specifically tailored for $\langle s,p,* \rangle$ questions. While Graph-CoT is generally applicable to any question type due to the reasoning flexibility of LLMs, its win rates are lower because the LLM often explores only a subset of relevant entities and derives answers from a limited view of the knowledge graph. This results in less comprehensive and diverse responses compared to methods that retrieve all relevant entities. However, this narrow focus allows Graph-CoT to perform comparably to \name{} in terms of directness.

In terms of execution efficiency and cost (see \autoref{tab:main_results}), \name{} achieves superior generation quality without a significant increase in response latency or token usage. Instead, it maintains relatively low latency and substantially reduces token consumption compared to more resource‐intensive baselines. Specifically, \name{} consistently records lower latency and requires fewer tokens than Fast-GraphRAG and Graph-CoT. This efficiency arises because Fast-GraphRAG relies on dynamic computation of Personalized PageRank (PPR) scores for each node during retrieval, and Graph-CoT invokes the LLM at every traversal step—both of which incur high computational complexity and resource usage. MS\_GraphRAG also exhibits high token consumption by retrieving large amounts of irrelevant information, even when questions target specific aspects. Although RoG and Cypher offer shorter latency and lower token usage, they do not match \name{}’s generation quality, which remains the primary criterion for GraphRAG applications.

\begin{table}[!t]
\centering
\caption{Win rates for each of the basic question patterns and nested questions.}
\label{tab:overall_decompose}
\resizebox{0.95\textwidth}{!}{%
\begin{tabular}{@{}ccccccc@{}}
\toprule
\textbf{Question Pattern} & \textbf{MS\_GraphRAG} & \textbf{RoG} & \textbf{Fast-graphrag} & \textbf{Graph-CoT} & \textbf{Cypher} & \textbf{PolyG} \\ \midrule
$\langle s,*,* \rangle$ & 30.00\% & 23.75\% & 2.92\% & 26.25\% & 15.00\% & \textbf{33.33\%} \\
$\langle s,p,* \rangle$ & 10.83\% & \textbf{47.08\%} & 24.58\% & 18.33\% & 13.33\% & 45.00\% \\
$\langle s,*,o \rangle$ & 25.83\% & 9.17\% & 5.42\% & 10.00\% & 3.75\% & \textbf{59.58\%} \\
$\langle s,p,o \rangle$ & 4.17\% & 1.25\% & 20.42\% & 5.83\% & 3.75\% & \textbf{70.42\%} \\
Nested & 15.00\% & 15.00\% & 7.92\% & 8.75\% & 3.75\% & \textbf{65.42\%} \\ \bottomrule
\end{tabular}%
}
\end{table}

\stitle{Response quality for different question patterns.} In~\autoref{tab:overall_decompose}, we break down the win rate by the four basic question patterns and nested questions. We observe that some baselines achieve win rates comparable to \name{} on $\langle s, *, * \rangle$ (e.g., MS\_GraphRAG) and $\langle s, p, * \rangle$ (e.g., RoG) questions. For the $\langle s, *, * \rangle$ pattern, questions are typically simple and seek general information about entities; MS\_GraphRAG is specifically designed and well-optimized for such tasks. RoG achieves similar win rates to \name{} on $\langle s, p, * \rangle$ questions because this pattern is the primary focus of existing KGQA benchmarks targeted by RoG. Consequently, similar performance trends are also observed on representative KGQA benchmarks, where \name{} matches RoG and outperforms all other baselines. Although specific baselines perform similarly to \name{} in terms of the “\textit{Overall Winner}” metric for these two question patterns, \name{} outperforms them in other criteria (e.g., \textit{Comprehensiveness}), making it a preferable choice for users with specific priorities. For all other patterns, \name{} constantly delivers superior response quality, achieving overall win rates above 60\%.

\begin{figure}[!t]
    \centering
    \begin{subfigure}[b]{0.48\textwidth}
        \centering
        \includegraphics[width=\textwidth]{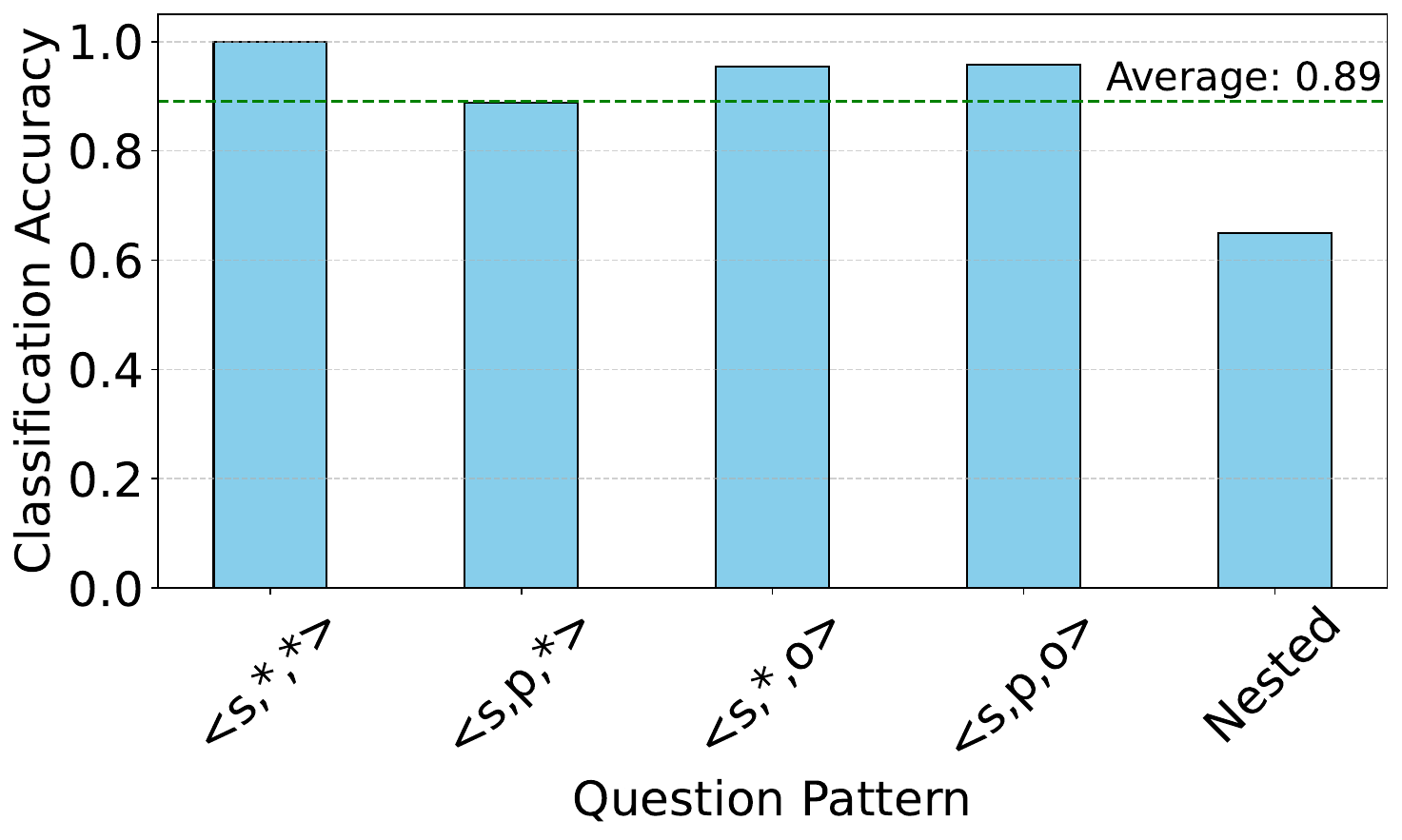}
        \caption{Accuracy of question categorization.}
        \label{fig:qc_acc}
    \end{subfigure}
    \begin{subfigure}[b]{0.48\textwidth}
        \centering
        \includegraphics[width=\textwidth]{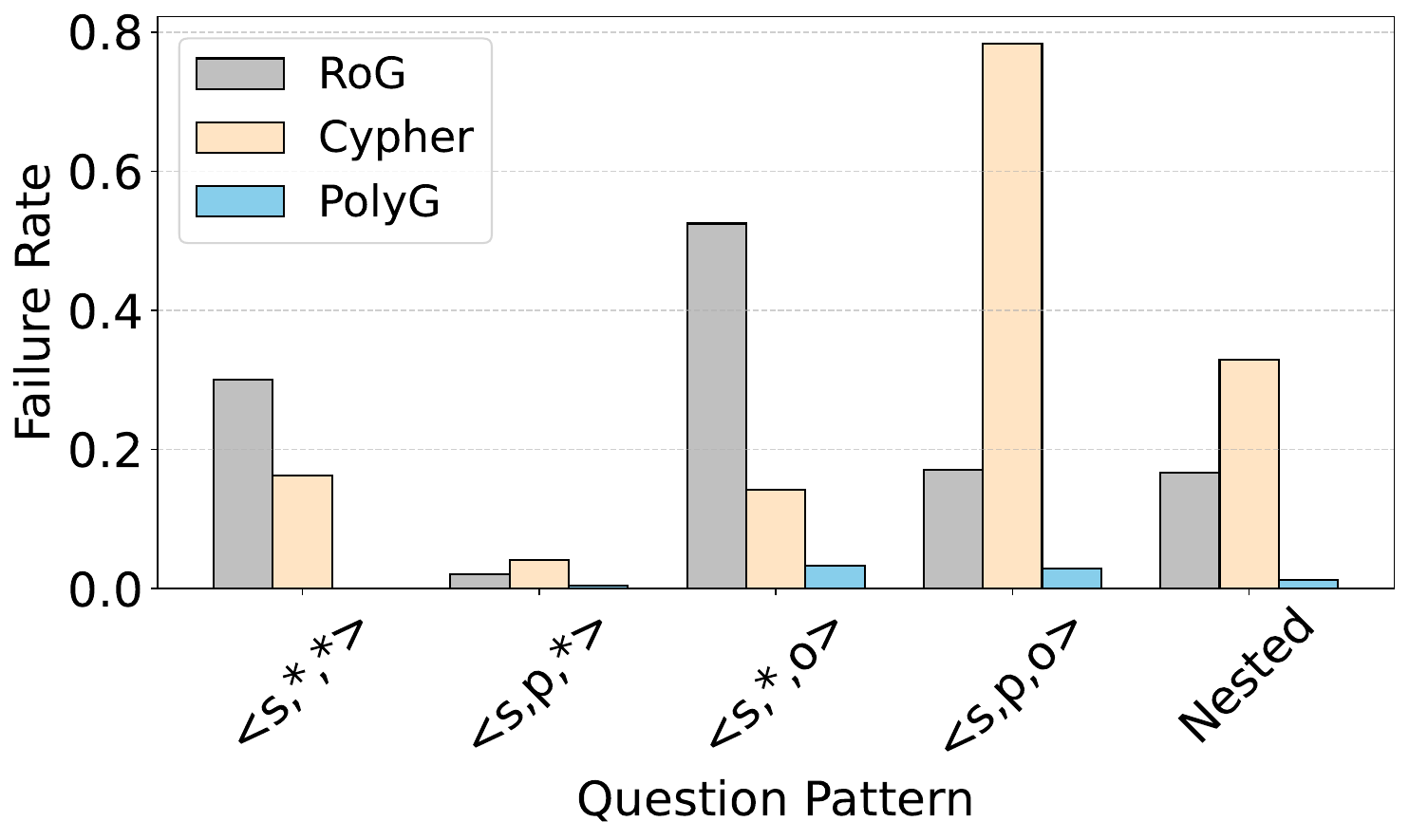}
        \caption{Failure rate comparison.}
        \label{fig:failure_rate}
    \end{subfigure}
    \caption{Question categorization accuracy and failure rates of \name{} on each question pattern.}
    \label{fig:micro_exp}
\end{figure}

\stitle{Accuracy of question categorization.} \autoref{fig:qc_acc} presents the accuracy of LLM-based question categorization in \name{} for different question patterns. The average accuracy is 89\%, demonstrating that our question pattern taxonomy is clear and effective for the LLM to differentiate between the patterns. Notably, accuracy is lower for $\langle s, p, * \rangle$ and nested questions compared to the other three patterns. This is because \name{} tends to decompose complex $\langle s, p, * \rangle$ questions involving multi-hop relations into multiple steps, while it merges simple nested questions of the form $\langle s, p, * \rangle + \langle s, p, * \rangle$ into a single $\langle s, p, * \rangle$ question. However, such decomposition and merging do not impair the LLM’s ability to generate quality responses, and thus we allow this intrinsic flexibility.

\stitle{Failure analysis.} Although the LLM is provided with the graph schema in \name{}, it can still generate incorrect Cypher queries—either semantically incorrect or returning empty results. To evaluate the effectiveness of our self-correction module during Cypher query generation, we report the failure rates broken down by the 4 basic and nested question patterns in~\autoref{fig:failure_rate}. For comparison, we also include the failure rates of RoG and Cypher. The results show that \name{} successfully generates correct Cypher queries for over 95\% of the questions, while RoG and Cypher fail for 24\% and 30\% of tasks, respectively. Furthermore, without the LLM self-correction module, \name{} would fail for 70\% of the nested questions. These findings suggest that our adaptive prompting and self-correction mechanism effectively reduce failures and enable more reliable Cypher query generation.

\stitle{Additional experiments.} In~\autoref{sec:result_decompose_appendix}, we provide additional experimental results. These include win rate decompositions across the other four criteria, detailed performance and efficiency comparisons on each dataset, and further analysis of the nested question pattern.

\vspace{-2mm}
\section{Conclusions}\label{sec:conclusion}

In this work, we develop a new benchmark, \benchmarkname{}, for general GraphRAG workloads, covering a wide range of graph question patterns to enable more reliable evaluation of GraphRAG methods. Furthermore, we propose a novel GraphRAG solution, \name{}, which excels at providing high-quality responses to diverse GraphRAG questions. \name{} is an adaptive framework that automatically identifies the question pattern and dynamically prompts the LLM to generate appropriate Cypher queries to retrieve the relevant context from the knowledge graphs. Experiment results show that \name{} achieves high response quality with a low response latency and token usage.

\bibliography{reference}

\begin{thebibliography}{41}
\providecommand{\natexlab}[1]{#1}
\providecommand{\url}[1]{\texttt{#1}}
\expandafter\ifx\csname urlstyle\endcsname\relax
  \providecommand{\doi}[1]{doi: #1}\else
  \providecommand{\doi}{doi: \begingroup \urlstyle{rm}\Url}\fi

\bibitem[Touvron et~al.(2023)Touvron, Martin, Stone, Albert, Almahairi, Babaei, Bashlykov, Batra, Bhargava, Bhosale, Bikel, Blecher, Ferrer, Chen, Cucurull, Esiobu, Fernandes, Fu, Fu, Fuller, Gao, Goswami, Goyal, Hartshorn, Hosseini, Hou, Inan, Kardas, Kerkez, Khabsa, Kloumann, Korenev, Koura, Lachaux, Lavril, Lee, Liskovich, Lu, Mao, Martinet, Mihaylov, Mishra, Molybog, Nie, Poulton, Reizenstein, Rungta, Saladi, Schelten, Silva, Smith, Subramanian, Tan, Tang, Taylor, Williams, Kuan, Xu, Yan, Zarov, Zhang, Fan, Kambadur, Narang, Rodriguez, Stojnic, Edunov, and Scialom]{llama2}
Hugo Touvron, Louis Martin, Kevin Stone, Peter Albert, Amjad Almahairi, Yasmine Babaei, Nikolay Bashlykov, Soumya Batra, Prajjwal Bhargava, Shruti Bhosale, Dan Bikel, Lukas Blecher, Cristian~Canton Ferrer, Moya Chen, Guillem Cucurull, David Esiobu, Jude Fernandes, Jeremy Fu, Wenyin Fu, Brian Fuller, Cynthia Gao, Vedanuj Goswami, Naman Goyal, Anthony Hartshorn, Saghar Hosseini, Rui Hou, Hakan Inan, Marcin Kardas, Viktor Kerkez, Madian Khabsa, Isabel Kloumann, Artem Korenev, Punit~Singh Koura, Marie-Anne Lachaux, Thibaut Lavril, Jenya Lee, Diana Liskovich, Yinghai Lu, Yuning Mao, Xavier Martinet, Todor Mihaylov, Pushkar Mishra, Igor Molybog, Yixin Nie, Andrew Poulton, Jeremy Reizenstein, Rashi Rungta, Kalyan Saladi, Alan Schelten, Ruan Silva, Eric~Michael Smith, Ranjan Subramanian, Xiaoqing~Ellen Tan, Binh Tang, Ross Taylor, Adina Williams, Jian~Xiang Kuan, Puxin Xu, Zheng Yan, Iliyan Zarov, Yuchen Zhang, Angela Fan, Melanie Kambadur, Sharan Narang, Aurelien Rodriguez, Robert Stojnic, Sergey Edunov, and Thomas Scialom.
\newblock Llama 2: Open foundation and fine-tuned chat models, 2023.
\newblock URL \url{https://arxiv.org/abs/2307.09288}.

\bibitem[OpenAI et~al.(2024)OpenAI, Achiam, Adler, Agarwal, Ahmad, Akkaya, Aleman, Almeida, Altenschmidt, and et~al.]{openai2024gpt4technicalreport}
OpenAI, Josh Achiam, Steven Adler, Sandhini Agarwal, Lama Ahmad, Ilge Akkaya, Florencia~Leoni Aleman, Diogo Almeida, Janko Altenschmidt, and Sam~Altman et~al.
\newblock Gpt-4 technical report, 2024.
\newblock URL \url{https://arxiv.org/abs/2303.08774}.

\bibitem[Brown et~al.(2020)Brown, Mann, Ryder, Subbiah, Kaplan, Dhariwal, Neelakantan, Shyam, Sastry, Askell, Agarwal, Herbert-Voss, Krueger, Henighan, Child, Ramesh, Ziegler, Wu, Winter, Hesse, Chen, Sigler, Litwin, Gray, Chess, Clark, Berner, McCandlish, Radford, Sutskever, and Amodei]{brown2020languagemodelsfewshotlearners}
Tom~B. Brown, Benjamin Mann, Nick Ryder, Melanie Subbiah, Jared Kaplan, Prafulla Dhariwal, Arvind Neelakantan, Pranav Shyam, Girish Sastry, Amanda Askell, Sandhini Agarwal, Ariel Herbert-Voss, Gretchen Krueger, Tom Henighan, Rewon Child, Aditya Ramesh, Daniel~M. Ziegler, Jeffrey Wu, Clemens Winter, Christopher Hesse, Mark Chen, Eric Sigler, Mateusz Litwin, Scott Gray, Benjamin Chess, Jack Clark, Christopher Berner, Sam McCandlish, Alec Radford, Ilya Sutskever, and Dario Amodei.
\newblock Language models are few-shot learners, 2020.
\newblock URL \url{https://arxiv.org/abs/2005.14165}.

\bibitem[Bang et~al.(2023)Bang, Cahyawijaya, Lee, Dai, Su, Wilie, Lovenia, Ji, Yu, Chung, Do, Xu, and Fung]{multitask}
Yejin Bang, Samuel Cahyawijaya, Nayeon Lee, Wenliang Dai, Dan Su, Bryan Wilie, Holy Lovenia, Ziwei Ji, Tiezheng Yu, Willy Chung, Quyet~V. Do, Yan Xu, and Pascale Fung.
\newblock A multitask, multilingual, multimodal evaluation of {C}hat{GPT} on reasoning, hallucination, and interactivity.
\newblock In Jong~C. Park, Yuki Arase, Baotian Hu, Wei Lu, Derry Wijaya, Ayu Purwarianti, and Adila~Alfa Krisnadhi, editors, \emph{Proceedings of the 13th International Joint Conference on Natural Language Processing and the 3rd Conference of the Asia-Pacific Chapter of the Association for Computational Linguistics (Volume 1: Long Papers)}, pages 675--718, Nusa Dua, Bali, November 2023. Association for Computational Linguistics.
\newblock \doi{10.18653/v1/2023.ijcnlp-main.45}.
\newblock URL \url{https://aclanthology.org/2023.ijcnlp-main.45/}.

\bibitem[Dam et~al.(2024)Dam, Hong, Qiao, and Zhang]{chatbot}
Sumit~Kumar Dam, Choong~Seon Hong, Yu~Qiao, and Chaoning Zhang.
\newblock A complete survey on llm-based ai chatbots, 2024.
\newblock URL \url{https://arxiv.org/abs/2406.16937}.

\bibitem[Wang et~al.(2023)Wang, Lyu, Ji, Zhang, Yu, Shi, and Tu]{translation}
Longyue Wang, Chenyang Lyu, Tianbo Ji, Zhirui Zhang, Dian Yu, Shuming Shi, and Zhaopeng Tu.
\newblock Document-level machine translation with large language models.
\newblock In Houda Bouamor, Juan Pino, and Kalika Bali, editors, \emph{Proceedings of the 2023 Conference on Empirical Methods in Natural Language Processing}, pages 16646--16661, Singapore, December 2023. Association for Computational Linguistics.
\newblock \doi{10.18653/v1/2023.emnlp-main.1036}.
\newblock URL \url{https://aclanthology.org/2023.emnlp-main.1036/}.

\bibitem[Sahoo et~al.(2024)Sahoo, Meharia, Ghosh, Saha, Jain, and Chadha]{hallucination}
Pranab Sahoo, Prabhash Meharia, Akash Ghosh, Sriparna Saha, Vinija Jain, and Aman Chadha.
\newblock A comprehensive survey of hallucination in large language, image, video and audio foundation models.
\newblock In Yaser Al-Onaizan, Mohit Bansal, and Yun-Nung Chen, editors, \emph{Findings of the Association for Computational Linguistics: EMNLP 2024}, pages 11709--11724, Miami, Florida, USA, November 2024. Association for Computational Linguistics.
\newblock \doi{10.18653/v1/2024.findings-emnlp.685}.
\newblock URL \url{https://aclanthology.org/2024.findings-emnlp.685/}.

\bibitem[Zhang et~al.(2023)Zhang, Li, Cui, Cai, Liu, Fu, Huang, Zhao, Zhang, Chen, Wang, Luu, Bi, Shi, and Shi]{zhang2023sirenssongaiocean}
Yue Zhang, Yafu Li, Leyang Cui, Deng Cai, Lemao Liu, Tingchen Fu, Xinting Huang, Enbo Zhao, Yu~Zhang, Yulong Chen, Longyue Wang, Anh~Tuan Luu, Wei Bi, Freda Shi, and Shuming Shi.
\newblock Siren's song in the ai ocean: A survey on hallucination in large language models, 2023.
\newblock URL \url{https://arxiv.org/abs/2309.01219}.

\bibitem[Edge et~al.(2024)Edge, Trinh, Cheng, Bradley, Chao, Mody, Truitt, and Larson]{ms_graphrag}
Darren Edge, Ha~Trinh, Newman Cheng, Joshua Bradley, Alex Chao, Apurva Mody, Steven Truitt, and Jonathan Larson.
\newblock From local to global: A graph rag approach to query-focused summarization, 2024.
\newblock URL \url{https://arxiv.org/abs/2404.16130}.

\bibitem[Wang et~al.(2018)Wang, Huang, Zhao, Zhang, Zhao, and Lee]{ecommerce}
Jizhe Wang, Pipei Huang, Huan Zhao, Zhibo Zhang, Binqiang Zhao, and Dik~Lun Lee.
\newblock Billion-scale commodity embedding for e-commerce recommendation in alibaba.
\newblock In \emph{KDD}, page 839–848, 2018.

\bibitem[Weber et~al.(2019)Weber, Domeniconi, Chen, Weidele, Bellei, Robinson, and Leiserson]{finance}
Mark Weber, Giacomo Domeniconi, Jie Chen, Daniel Karl~I. Weidele, Claudio Bellei, Tom Robinson, and Charles~E. Leiserson.
\newblock Anti-money laundering in bitcoin: Experimenting with graph convolutional networks for financial forensics.
\newblock \emph{CoRR}, 2019.

\bibitem[Garton et~al.(1997)Garton, Haythornthwaite, and Wellman]{social}
Laura Garton, Caroline Haythornthwaite, and Barry Wellman.
\newblock Studying online social networks.
\newblock \emph{J. Comput. Mediat. Commun.}, 1997.

\bibitem[Guo et~al.(2024)Guo, Xia, Yu, Ao, and Huang]{lightrag}
Zirui Guo, Lianghao Xia, Yanhua Yu, Tu~Ao, and Chao Huang.
\newblock Lightrag: Simple and fast retrieval-augmented generation, 2024.
\newblock URL \url{https://arxiv.org/abs/2410.05779}.

\bibitem[Luo et~al.(2024)Luo, Li, Haf, and Pan]{rog}
Linhao Luo, Yuan-Fang Li, Reza Haf, and Shirui Pan.
\newblock Reasoning on graphs: Faithful and interpretable large language model reasoning.
\newblock In \emph{The Twelfth International Conference on Learning Representations}, 2024.
\newblock URL \url{https://openreview.net/forum?id=ZGNWW7xZ6Q}.

\bibitem[fas(2024)]{fast_graphrag}
Streamlined and promptable fast graphrag framework designed for interpretable, high-precision, agent-driven retrieval workflows., 2024.
\newblock URL \url{https://github.com/circlemind-ai/fast-graphrag}.

\bibitem[Gutierrez et~al.(2024)Gutierrez, Shu, Gu, Yasunaga, and Su]{hipporag}
Bernal~Jimenez Gutierrez, Yiheng Shu, Yu~Gu, Michihiro Yasunaga, and Yu~Su.
\newblock Hippo{RAG}: Neurobiologically inspired long-term memory for large language models.
\newblock In \emph{The Thirty-eighth Annual Conference on Neural Information Processing Systems}, 2024.
\newblock URL \url{https://openreview.net/forum?id=hkujvAPVsg}.

\bibitem[Page et~al.(1999)Page, Brin, Motwani, and Winograd]{ppr}
Lawrence Page, Sergey Brin, Rajeev Motwani, and Terry Winograd.
\newblock The pagerank citation ranking: Bringing order to the web.
\newblock Technical Report 1999-66, Stanford InfoLab, November 1999.
\newblock URL \url{http://ilpubs.stanford.edu:8090/422/}.
\newblock Previous number = SIDL-WP-1999-0120.

\bibitem[Sun et~al.(2024)Sun, Xu, Tang, Wang, Lin, Gong, Ni, Shum, and Guo]{tog}
Jiashuo Sun, Chengjin Xu, Lumingyuan Tang, Saizhuo Wang, Chen Lin, Yeyun Gong, Lionel Ni, Heung-Yeung Shum, and Jian Guo.
\newblock Think-on-graph: Deep and responsible reasoning of large language model on knowledge graph.
\newblock In \emph{The Twelfth International Conference on Learning Representations}, 2024.
\newblock URL \url{https://openreview.net/forum?id=nnVO1PvbTv}.

\bibitem[Jin et~al.(2024)Jin, Xie, Zhang, Roy, Zhang, Li, Li, Tang, Wang, Meng, and Han]{graphcot}
Bowen Jin, Chulin Xie, Jiawei Zhang, {Kashob Kumar} Roy, Yu~Zhang, Zheng Li, Ruirui Li, Xianfeng Tang, Suhang Wang, Yu~Meng, and Jiawei Han.
\newblock Graph chain-of-thought: Augmenting large language models by reasoning on graphs.
\newblock In Lun-Wei Ku, Andre Martins, and Vivek Srikumar, editors, \emph{62nd Annual Meeting of the Association for Computational Linguistics, ACL 2024 - Proceedings of the Conference}, Proceedings of the Annual Meeting of the Association for Computational Linguistics, pages 163--184. Association for Computational Linguistics (ACL), 2024.

\bibitem[He et~al.(2024)He, Tian, Sun, Chawla, Laurent, LeCun, Bresson, and Hooi]{gretriever}
Xiaoxin He, Yijun Tian, Yifei Sun, Nitesh~V Chawla, Thomas Laurent, Yann LeCun, Xavier Bresson, and Bryan Hooi.
\newblock G-retriever: Retrieval-augmented generation for textual graph understanding and question answering.
\newblock In \emph{The Thirty-eighth Annual Conference on Neural Information Processing Systems}, 2024.
\newblock URL \url{https://openreview.net/forum?id=MPJ3oXtTZl}.

\bibitem[Mavromatis and Karypis(2025)]{gnnrag}
Costas Mavromatis and George Karypis.
\newblock {GNN}-{RAG}: Graph neural retrieval for large language model reasoning, 2025.
\newblock URL \url{https://openreview.net/forum?id=EVuANndPlX}.

\bibitem[Li et~al.(2025)Li, Miao, and Li]{subgraphrag}
Mufei Li, Siqi Miao, and Pan Li.
\newblock Simple is effective: The roles of graphs and large language models in knowledge-graph-based retrieval-augmented generation.
\newblock In \emph{ICLR 2025 Workshop on Foundation Models in the Wild}, 2025.
\newblock URL \url{https://openreview.net/forum?id=2NbxnNI94F}.

\bibitem[Bordes et~al.(2015)Bordes, Usunier, Chopra, and Weston]{simpleq}
Antoine Bordes, Nicolas Usunier, Sumit Chopra, and Jason Weston.
\newblock Large-scale simple question answering with memory networks.
\newblock \emph{CoRR}, abs/1506.02075, 2015.
\newblock URL \url{http://arxiv.org/abs/1506.02075}.

\bibitem[Yih et~al.(2016)Yih, Richardson, Meek, Chang, and Suh]{webqsp}
Wen-tau Yih, Matthew Richardson, Chris Meek, Ming-Wei Chang, and Jina Suh.
\newblock The value of semantic parse labeling for knowledge base question answering.
\newblock In Katrin Erk and Noah~A. Smith, editors, \emph{Proceedings of the 54th Annual Meeting of the Association for Computational Linguistics (Volume 2: Short Papers)}, pages 201--206, Berlin, Germany, August 2016. Association for Computational Linguistics.
\newblock \doi{10.18653/v1/P16-2033}.
\newblock URL \url{https://aclanthology.org/P16-2033/}.

\bibitem[Su et~al.(2016)Su, Sun, Sadler, Srivatsa, G{\"u}r, Yan, and Yan]{graphq}
Yu~Su, Huan Sun, Brian Sadler, Mudhakar Srivatsa, Izzeddin G{\"u}r, Zenghui Yan, and Xifeng Yan.
\newblock On generating characteristic-rich question sets for {QA} evaluation.
\newblock In Jian Su, Kevin Duh, and Xavier Carreras, editors, \emph{Proceedings of the 2016 Conference on Empirical Methods in Natural Language Processing}, pages 562--572, Austin, Texas, November 2016. Association for Computational Linguistics.
\newblock \doi{10.18653/v1/D16-1054}.
\newblock URL \url{https://aclanthology.org/D16-1054/}.

\bibitem[Talmor and Berant(2018)]{cwq}
Alon Talmor and Jonathan Berant.
\newblock The web as a knowledge-base for answering complex questions.
\newblock In Marilyn Walker, Heng Ji, and Amanda Stent, editors, \emph{Proceedings of the 2018 Conference of the North {A}merican Chapter of the Association for Computational Linguistics: Human Language Technologies, Volume 1 (Long Papers)}, pages 641--651, New Orleans, Louisiana, June 2018. Association for Computational Linguistics.
\newblock \doi{10.18653/v1/N18-1059}.
\newblock URL \url{https://aclanthology.org/N18-1059/}.

\bibitem[Gu et~al.(2021)Gu, Kase, Vanni, Sadler, Liang, Yan, and Su]{grailqa}
Yu~Gu, Sue Kase, Michelle Vanni, Brian Sadler, Percy Liang, Xifeng Yan, and Yu~Su.
\newblock Beyond i.i.d.: Three levels of generalization for question answering on knowledge bases.
\newblock In \emph{Proceedings of the Web Conference 2021}, WWW '21, page 3477–3488, New York, NY, USA, 2021. Association for Computing Machinery.
\newblock ISBN 9781450383127.
\newblock \doi{10.1145/3442381.3449992}.
\newblock URL \url{https://doi.org/10.1145/3442381.3449992}.

\bibitem[Zhang et~al.(2025)Zhang, Perry, Dulepet, Ji, Menders, Lin, Jones, Hussein, Liu, Jasper, Peetathawatchai, Glenn, Sivashankar, Zamoshchin, Glikbarg, Askaryar, Yang, Zhang, Alluri, Tran, Sangpisit, Yiorkadjis, Osele, Raghupathi, Boneh, Ho, and Liang]{cypherbench}
Andy~K. Zhang, Neil Perry, Riya Dulepet, Joey Ji, Celeste Menders, Justin~W. Lin, Eliot Jones, Gashon Hussein, Samantha Liu, Donovan Jasper, Pura Peetathawatchai, Ari Glenn, Vikram Sivashankar, Daniel Zamoshchin, Leo Glikbarg, Derek Askaryar, Mike Yang, Teddy Zhang, Rishi Alluri, Nathan Tran, Rinnara Sangpisit, Polycarpos Yiorkadjis, Kenny Osele, Gautham Raghupathi, Dan Boneh, Daniel~E. Ho, and Percy Liang.
\newblock Cybench: A framework for evaluating cybersecurity capabilities and risks of language models, 2025.
\newblock URL \url{https://arxiv.org/abs/2408.08926}.

\bibitem[Yang et~al.(2015)Yang, Yih, and Meek]{wikiqa}
Yi~Yang, Wen-tau Yih, and Christopher Meek.
\newblock {W}iki{QA}: A challenge dataset for open-domain question answering.
\newblock In Llu{\'i}s M{\`a}rquez, Chris Callison-Burch, and Jian Su, editors, \emph{Proceedings of the 2015 Conference on Empirical Methods in Natural Language Processing}, pages 2013--2018, Lisbon, Portugal, September 2015. Association for Computational Linguistics.
\newblock \doi{10.18653/v1/D15-1237}.
\newblock URL \url{https://aclanthology.org/D15-1237/}.

\bibitem[Francis et~al.(2018)Francis, Green, Guagliardo, Libkin, Lindaaker, Marsault, Plantikow, Rydberg, Selmer, and Taylor]{cypher}
Nadime Francis, Alastair Green, Paolo Guagliardo, Leonid Libkin, Tobias Lindaaker, Victor Marsault, Stefan Plantikow, Mats Rydberg, Petra Selmer, and Andr\'{e}s Taylor.
\newblock Cypher: An evolving query language for property graphs.
\newblock In \emph{Proceedings of the 2018 International Conference on Management of Data}, SIGMOD '18, page 1433–1445, New York, NY, USA, 2018. Association for Computing Machinery.
\newblock ISBN 9781450347037.
\newblock \doi{10.1145/3183713.3190657}.
\newblock URL \url{https://doi.org/10.1145/3183713.3190657}.

\bibitem[AOL(2025)]{aolquerylog}
AOL, 2025.
\newblock URL \url{https://techcrunch.com/2006/08/07/aol-this-was-a-screw-up/}.

\bibitem[Leipzig(2022)]{academia}
Database~Group Leipzig.
\newblock Dblp-scholar, 2022.
\newblock URL \url{https://dbs.uni-leipzig.de/file/DBLP-Scholar.zip}.

\bibitem[Graph(2022)]{literature}
UCSD~Book Graph.
\newblock Goodreads, 2022.
\newblock URL \url{https://sites.google.com/eng.ucsd.edu/ucsdbookgraph/home}.

\bibitem[He and McAuley(2016)]{e-commerce}
Ruining He and Julian McAuley.
\newblock Ups and downs: Modeling the visual evolution of fashion trends with one-class collaborative filtering.
\newblock In \emph{Proceedings of the 25th International Conference on World Wide Web}, WWW '16, page 507–517, Republic and Canton of Geneva, CHE, 2016. International World Wide Web Conferences Steering Committee.
\newblock ISBN 9781450341431.
\newblock \doi{10.1145/2872427.2883037}.
\newblock URL \url{https://doi.org/10.1145/2872427.2883037}.

\bibitem[Anthropic(2024)]{claude}
Anthropic.
\newblock Claude 3.5 sonnet, 2024.
\newblock URL \url{https://www.anthropic.com/}.

\bibitem[Angles et~al.(2020)Angles, Thakkar, and Tomaszuk]{rdf2property}
Renzo Angles, Harsh Thakkar, and Dominik Tomaszuk.
\newblock Mapping rdf databases to property graph databases.
\newblock \emph{IEEE Access}, 8:\penalty0 86091--86110, 2020.
\newblock \doi{10.1109/ACCESS.2020.2993117}.

\bibitem[Perozzi et~al.(2014)Perozzi, Al-Rfou, and Skiena]{deepwalk}
Bryan Perozzi, Rami Al-Rfou, and Steven Skiena.
\newblock Deepwalk: online learning of social representations.
\newblock In \emph{Proceedings of the 20th ACM SIGKDD International Conference on Knowledge Discovery and Data Mining}, KDD '14, page 701–710, New York, NY, USA, 2014. Association for Computing Machinery.
\newblock ISBN 9781450329569.
\newblock \doi{10.1145/2623330.2623732}.
\newblock URL \url{https://doi.org/10.1145/2623330.2623732}.

\bibitem[Grover and Leskovec(2016)]{node2vec}
Aditya Grover and Jure Leskovec.
\newblock node2vec: Scalable feature learning for networks.
\newblock In \emph{Proceedings of the 22nd ACM SIGKDD International Conference on Knowledge Discovery and Data Mining}, KDD '16, page 855–864, New York, NY, USA, 2016. Association for Computing Machinery.
\newblock ISBN 9781450342322.
\newblock \doi{10.1145/2939672.2939754}.
\newblock URL \url{https://doi.org/10.1145/2939672.2939754}.

\bibitem[DeepSeek-AI et~al.(2025)DeepSeek-AI, Guo, Yang, Zhang, Song, Zhang, Xu, Zhu, Ma, Wang, Bi, Zhang, Yu, Wu, Wu, Gou, Shao, Li, Gao, Liu, Xue, Wang, Wu, Feng, Lu, Zhao, Deng, Zhang, Ruan, Dai, Chen, Ji, Li, Lin, Dai, Luo, Hao, Chen, Li, Zhang, Bao, Xu, Wang, Ding, Xin, Gao, Qu, Li, Guo, Li, Wang, Chen, Yuan, Qiu, Li, Cai, Ni, Liang, Chen, Dong, Hu, Gao, Guan, Huang, Yu, Wang, Zhang, Zhao, Wang, Zhang, Xu, Xia, Zhang, Zhang, Tang, Li, Wang, Li, Tian, Huang, Zhang, Wang, Chen, Du, Ge, Zhang, Pan, Wang, Chen, Jin, Chen, Lu, Zhou, Chen, Ye, Wang, Yu, Zhou, Pan, Li, Zhou, Wu, Ye, Yun, Pei, Sun, Wang, Zeng, Zhao, Liu, Liang, Gao, Yu, Zhang, Xiao, An, Liu, Wang, Chen, Nie, Cheng, Liu, Xie, Liu, Yang, Li, Su, Lin, Li, Jin, Shen, Chen, Sun, Wang, Song, Zhou, Wang, Shan, Li, Wang, Wei, Zhang, Xu, Li, Zhao, Sun, Wang, Yu, Zhang, Shi, Xiong, He, Piao, Wang, Tan, Ma, Liu, Guo, Ou, Wang, Gong, Zou, He, Xiong, Luo, You, Liu, Zhou, Zhu, Xu, Huang, Li, Zheng, Zhu, Ma, Tang, Zha, Yan, Ren, Ren, Sha, Fu, Xu, Xie, Zhang, Hao, Ma, Yan, Wu, Gu, Zhu, Liu, Li, Xie, Song, Pan, Huang, Xu, Zhang, and Zhang]{deepseekr1}
DeepSeek-AI, Daya Guo, Dejian Yang, Haowei Zhang, Junxiao Song, Ruoyu Zhang, Runxin Xu, Qihao Zhu, Shirong Ma, Peiyi Wang, Xiao Bi, Xiaokang Zhang, Xingkai Yu, Yu~Wu, Z.~F. Wu, Zhibin Gou, Zhihong Shao, Zhuoshu Li, Ziyi Gao, Aixin Liu, Bing Xue, Bingxuan Wang, Bochao Wu, Bei Feng, Chengda Lu, Chenggang Zhao, Chengqi Deng, Chenyu Zhang, Chong Ruan, Damai Dai, Deli Chen, Dongjie Ji, Erhang Li, Fangyun Lin, Fucong Dai, Fuli Luo, Guangbo Hao, Guanting Chen, Guowei Li, H.~Zhang, Han Bao, Hanwei Xu, Haocheng Wang, Honghui Ding, Huajian Xin, Huazuo Gao, Hui Qu, Hui Li, Jianzhong Guo, Jiashi Li, Jiawei Wang, Jingchang Chen, Jingyang Yuan, Junjie Qiu, Junlong Li, J.~L. Cai, Jiaqi Ni, Jian Liang, Jin Chen, Kai Dong, Kai Hu, Kaige Gao, Kang Guan, Kexin Huang, Kuai Yu, Lean Wang, Lecong Zhang, Liang Zhao, Litong Wang, Liyue Zhang, Lei Xu, Leyi Xia, Mingchuan Zhang, Minghua Zhang, Minghui Tang, Meng Li, Miaojun Wang, Mingming Li, Ning Tian, Panpan Huang, Peng Zhang, Qiancheng Wang, Qinyu Chen, Qiushi Du, Ruiqi Ge, Ruisong Zhang, Ruizhe Pan, Runji Wang, R.~J. Chen, R.~L. Jin, Ruyi Chen, Shanghao Lu, Shangyan Zhou, Shanhuang Chen, Shengfeng Ye, Shiyu Wang, Shuiping Yu, Shunfeng Zhou, Shuting Pan, S.~S. Li, Shuang Zhou, Shaoqing Wu, Shengfeng Ye, Tao Yun, Tian Pei, Tianyu Sun, T.~Wang, Wangding Zeng, Wanjia Zhao, Wen Liu, Wenfeng Liang, Wenjun Gao, Wenqin Yu, Wentao Zhang, W.~L. Xiao, Wei An, Xiaodong Liu, Xiaohan Wang, Xiaokang Chen, Xiaotao Nie, Xin Cheng, Xin Liu, Xin Xie, Xingchao Liu, Xinyu Yang, Xinyuan Li, Xuecheng Su, Xuheng Lin, X.~Q. Li, Xiangyue Jin, Xiaojin Shen, Xiaosha Chen, Xiaowen Sun, Xiaoxiang Wang, Xinnan Song, Xinyi Zhou, Xianzu Wang, Xinxia Shan, Y.~K. Li, Y.~Q. Wang, Y.~X. Wei, Yang Zhang, Yanhong Xu, Yao Li, Yao Zhao, Yaofeng Sun, Yaohui Wang, Yi~Yu, Yichao Zhang, Yifan Shi, Yiliang Xiong, Ying He, Yishi Piao, Yisong Wang, Yixuan Tan, Yiyang Ma, Yiyuan Liu, Yongqiang Guo, Yuan Ou, Yuduan Wang, Yue Gong, Yuheng Zou, Yujia He, Yunfan Xiong, Yuxiang Luo, Yuxiang You, Yuxuan Liu, Yuyang Zhou, Y.~X. Zhu, Yanhong Xu, Yanping Huang, Yaohui Li, Yi~Zheng, Yuchen Zhu, Yunxian Ma, Ying Tang, Yukun Zha, Yuting Yan, Z.~Z. Ren, Zehui Ren, Zhangli Sha, Zhe Fu, Zhean Xu, Zhenda Xie, Zhengyan Zhang, Zhewen Hao, Zhicheng Ma, Zhigang Yan, Zhiyu Wu, Zihui Gu, Zijia Zhu, Zijun Liu, Zilin Li, Ziwei Xie, Ziyang Song, Zizheng Pan, Zhen Huang, Zhipeng Xu, Zhongyu Zhang, and Zhen Zhang.
\newblock Deepseek-r1: Incentivizing reasoning capability in llms via reinforcement learning, 2025.
\newblock URL \url{https://arxiv.org/abs/2501.12948}.

\bibitem[Berant et~al.(2013)Berant, Chou, Frostig, and Liang]{webquestion}
Jonathan Berant, Andrew Chou, Roy Frostig, and Percy Liang.
\newblock Semantic parsing on {F}reebase from question-answer pairs.
\newblock In David Yarowsky, Timothy Baldwin, Anna Korhonen, Karen Livescu, and Steven Bethard, editors, \emph{Proceedings of the 2013 Conference on Empirical Methods in Natural Language Processing}, pages 1533--1544, Seattle, Washington, USA, October 2013. Association for Computational Linguistics.
\newblock URL \url{https://aclanthology.org/D13-1160/}.

\bibitem[Google(2025)]{googlesuggestapi}
Google.
\newblock Google suggest api, 2025.
\newblock URL \url{https://developers.google.com/workspace/cloud-search/docs/reference/rest/v1/query/suggest}.

\end{thebibliography}

\newpage

\appendix
\section{Impact Statement}\label{sec:broader_impact}

While this work is focused solely on advancing the field of machine learning and is not expected to introduce any societal risks, we acknowledge that there are certainly cases where the large language model (LLM) applications can be possibly developed to cause unintended social implications, whether deliberately or inadvertently. For instance, the improper use of GraphRAG could inadvertently expose confidential user data from internal company databases to unauthorized parties.

\section{Detailed Experimental Settings}\label{sec:exp_settings_appendix}

\stitle{Hardware configurations.} We conduct all the experiments on an AWS EC2 g5.16xlarge instance with 1 NVIDIA A10 GPU (24GB memory), 64-core vCPUs, 256G memory and 2TB NVMe SSD storage. For LLM service, we use AWS bedrock API with Claude-3.5-sonnet~\cite{claude} and Deepseek-R1~\cite{deepseekr1} models.

\stitle{Baselines and models.} We compare \name{} against existing GraphRAG methods discussed in \autoref{sec:intro}. To ensure a fair and clear comparison, we categorize baselines based on their adopted graph traversal methods and select one representative from each category, despite minor differences in other modules. For instance, since both MS\_GraphRAG and LightRAG use BFS to retrieve all one-hop neighbors, we select MS\_GraphRAG as the representative method. Similarly, for traversal methods that invoke LLMs at each step, we compare with Graph-CoT instead of ToG. Additionally, since RoG requires fine-tuning an LLM to generate faithful reasoning paths, we instead prompt the LLM with the graph schema to avoid instruction tuning. For recent methods that require training of the retriever, such as G-retriever~\cite{gretriever} and SubgraphRAG~\cite{gnnrag}, we do not include them in this experiment since they can not be applied in general GraphRAG workloads where questions do not have ground-truth labels. We also include the baseline, named Cypher, which directly generates cypher queries using the LLM without the guidance of our proposed question pattern taxonomy and any further optimizations built upon it to show the necessity of our adaptive prompting. For \name{}, we set the $k$ in top-$k$ shortest paths to 10, which means that we only take the top-10 reasoning paths satisfying the optional constraints in the question. For consistency, we use the Claude-3.5-sonnet~\cite{claude} as the underlying LLM service for \name{} and all baselines, both for categorization and generation.

\stitle{Evaluation metric.} Since most of the graph question patterns do not have a single definitive answer as discussed in~\autoref{sec:question_categorization}, we assess the generation quality using win rates determined by LLM judgments. We evaluate responses from four perspectives: \textit{Comprehensiveness}, \textit{Diversity}, \textit{Empowerment}, and \textit{Directness}, and then let the LLM determines the \textit{Overall Winner} considering all criteria, following previous approaches~\cite{ms_graphrag, lightrag}. Full prompts for these criteria are provided in \autoref{sec:criteria_appendix}. As each question is judged across multiple baselines and thus there could be multiple high-qualty responses, we prompt the LLM to allow multiple winners for each criterion, leading to total win rates exceeding 100\%. In addition to win rates, we report \textit{F1-score} and \textit{Deepseek-R1} for $\langle s, p, * \rangle$ and $\langle s, p, * \rangle+ \langle s, p, * \rangle$ questions, which we have automatically labeled ground-truth answers. We use the strong open-weight reasoning model Deepseek-R1~\cite{deepseekr1} for judgment. Meanwhile, we measure end-to-end response latency and token usage to evaluate the computational efficiency.
\section{Additional Experimental Results}\label{sec:result_decompose_appendix}

\subsection{More Details on Generation Quality Results}

\begin{table}[t!]
\caption{Response generation quality on each basic question pattern and the nested question pattern.}
\label{tab:quality_decompose}
\resizebox{\textwidth}{!}{%
\begin{tabular}{@{}clcccccc@{}}
\toprule
\textbf{Question Pattern} & \textbf{Criteria} & \textbf{MS\_GraphRAG} & \textbf{RoG} & \textbf{Fast-graphrag} & \textbf{Graph-CoT} & \textbf{Cypher} & \textbf{PolyG} \\ \midrule
\multirow{5}{*}{$\langle s,*,* \rangle $} & Comprehensiveness & 52.50\% & 32.50\% & 5.83\% & 35.83\% & 33.33\% & \textbf{69.58\%} \\
 & Diversity & \textbf{43.33\%} & 27.08\% & 5.00\% & 40.42\% & 23.75\% & 42.08\% \\
 & Empowerment & 41.67\% & 35.83\% & 6.25\% & 31.25\% & 27.50\% & \textbf{48.75\%} \\
 & Directness & \textbf{54.17\%} & 19.58\% & 28.75\% & 32.50\% & 46.25\% & 48.75\% \\
 & Overall Winner & 30.00\% & 23.75\% & 2.92\% & 26.25\% & 15.00\% & \textbf{33.33\%} \\ \midrule
\multirow{8}{*}{$\langle s,p,* \rangle $} & Comprehensiveness & 17.08\% & 67.50\% & 33.75\% & 25.83\% & 17.50\% & \textbf{67.92\%} \\
 & Diversity & 5.42\% & 65.42\% & 17.08\% & 5.00\% & 7.92\% & \textbf{68.75\%} \\
 & Empowerment & 13.75\% & \textbf{59.58\%} & 27.50\% & 5.42\% & 18.33\% & \textbf{59.58\%} \\
 & Directness & 7.08\% & 29.58\% & 25.42\% & \textbf{56.67\%} & 9.58\% & 25.00\% \\
 & Overall Winner & 10.83\% & \textbf{47.08\%} & 24.58\% & 18.33\% & 13.33\% & 45.00\% \\ \cmidrule(l){2-8}
 & F1-score & 0.2224 & 0.7433 & 0.3659 & 0.4077 & 0.2182 & \textbf{0.7522} \\
 & Hit & 0.4625 & 0.9167 & 0.7208 & 0.7083 & 0.4083 & \textbf{0.9292} \\ \midrule
\multirow{5}{*}{$\langle s,*,o \rangle $} & Comprehensiveness & 32.08\% & 15.00\% & 7.50\% & 13.75\% & 6.25\% & \textbf{81.67\%} \\
& Diversity & 23.75\% & 11.25\% & 4.58\% & 17.08\% & 2.08\% & \textbf{80.00\%} \\
& Empowerment & 34.58\% & 13.33\% & 8.75\% & 13.75\% & 6.67\% & \textbf{71.67\%} \\
& Directness & \textbf{51.25\%} & 8.33\% & 37.92\% & 41.25\% & 10.83\% & 42.08\% \\
& Overall Winner & 25.83\% & 9.17\% & 5.42\% & 10.00\% & 3.75\% & \textbf{59.58\%} \\ \midrule
\multirow{5}{*}{$\langle s,p,o \rangle $} & Comprehensiveness & 7.92\% & 6.25\% & 26.25\% & 9.17\% & 5.00\% & \textbf{79.58\%} \\
 & Diversity & 6.25\% & 7.08\% & 19.58\% & 8.33\% & 4.58\% & \textbf{69.17\%} \\
 & Empowerment & 8.75\% & 4.58\% & 25.00\% & 6.67\% & 5.00\% & \textbf{74.58\%} \\
 & Directness & 4.58\% & 5.42\% & 35.42\% & 43.75\% & 4.58\% & \textbf{72.92\%} \\
 & Overall Winner & 4.17\% & 1.25\% & 20.42\% & 5.83\% & 3.75\% & \textbf{70.42\%} \\ \midrule
\multirow{8}{*}{Nested} & Comprehensiveness & 20.83\% & 22.50\% & 12.50\% & 12.50\% & 4.17\% & \textbf{74.58\%} \\
 & Diversity & 16.25\% & 20.83\% & 7.08\% & 8.33\% & 2.92\% & \textbf{70.42\%} \\
 & Empowerment & 23.33\% & 17.92\% & 11.67\% & 10.83\% & 5.83\% & \textbf{69.58\%} \\
 & Directness & 32.08\% & 15.83\% & 31.67\% & 37.92\% & 6.25\% & \textbf{44.58\%} \\
 & Overall Winner & 15.00\% & 15.00\% & 7.92\% & 8.75\% & 3.75\% & \textbf{65.42\%} \\ \cmidrule(l){2-8}
 & F1-score & 0.1573 & 0.4938 & 0.1884 & 0.3115 & 0.0022 & \textbf{0.5332} \\
 & Hit & 0.3500 & \textbf{0.7667} & 0.3667 & 0.4167 & 0.0167 & 0.7333 \\ \bottomrule
\end{tabular}%
}
\end{table}

\stitle{Decomposition over question patterns.} \autoref{tab:quality_decompose} compares the decomposed generation quality of \name{} and baseline methods with Claude-3.5-sonnet across all five evaluation criteria (averaged) and two quantitative metrics on all three knowledge graphs. The results show that \name{} achieves the highest win rates in 20 out of 25 cases, and also produces the highest \textit{F1-score} and \textit{Hit} on the basic $\langle s,p,* \rangle$ questions and the nested $\langle s,p,* \rangle + \langle s,p,* \rangle$ questions. Notably, \name{} outperforms all baselines across all criteria on the $\langle s,p,o \rangle$ and nested question types. For the remaining three patterns, \name{} has slightly lower win rates under the ``\textit{Directness}'' criterion. This is primarily because \name{} retrieves a more comprehensive context than the baselines, leading the LLM to generate longer responses that cover diverse aspects of the information. This observation is supported by the high win rates of \name{} in terms of ``\textit{Comprehensiveness},'' suggesting a trade-off between ``\textit{Directness}'' and ``\textit{Comprehensiveness}.'' Therefore, if users prioritize more comprehensive responses—or seek a method that performs robustly across various question patterns—\name{} is the preferable choice.

\begin{table}[!t]
\caption{Response generation quality of \name{} and baseline methods on each knowledge graph.}
\label{tab:quality_dataset}
\centering
\resizebox{\textwidth}{!}{%
\begin{tabular}{@{}clcccccc@{}}
\toprule
\textbf{Dataset} & \textbf{Criteria} & \textbf{MS\_GraphRGA} & \textbf{RoG} & \textbf{Fast-graphrag} & \textbf{Graph-CoT} & \textbf{Cypher} & \textbf{PolyG} \\ \midrule
\multirow{8}{*}{Academia} & Comprehensiveness & 29.00\% & 26.75\% & 12.00\% & 21.50\% & 10.75\% & \textbf{79.75\%} \\
 & Diversity & 22.50\% & 23.50\% & 7.75\% & 12.25\% & 5.00\% & \textbf{74.00\%} \\
 & Empowerment & 27.50\% & 22.50\% & 12.00\% & 12.50\% & 9.50\% & \textbf{71.00\%} \\
 & Directness & 31.00\% & 13.25\% & 29.75\% & \textbf{51.25\%} & 14.25\% & 48.25\% \\
 & Overall & 19.00\% & 17.00\% & 9.00\% & 14.75\% & 6.50\% & \textbf{61.50\%} \\ \cmidrule(l){2-8} 
 & F1-score & 0.2132 & 0.7551 & 0.371 & 0.5445 & 0.1332 & \textbf{0.7577} \\
 & Hit & 0.6000 & \textbf{0.9600} & 0.8400 & 0.8700 & 0.3600 & 0.9300 \\ \midrule
\multirow{8}{*}{Literature} & Comprehensiveness & 26.25\% & 27.50\% & 21.00\% & 19.25\% & 13.75\% & \textbf{73.50\%} \\
 & Diversity & 16.75\% & 26.50\% & 10.25\% & 15.50\% & 8.00\% & \textbf{64.75\%} \\
 & Empowerment & 23.25\% & 25.50\% & 18.25\% & 15.00\% & 12.00\% & \textbf{63.50\%} \\
 & Directness & 30.25\% & 12.50\% & 34.50\% & 38.00\% & 17.00\% & \textbf{48.75\%} \\
 & Overall & 16.75\% & 18.25\% & 13.25\% & 14.75\% & 8.50\% & \textbf{54.00\%} \\ \cmidrule(l){2-8} 
 & F1-score & 0.199 & 0.7558 & 0.3575 & 0.3422 & 0.2041 & \textbf{0.7788} \\
 & Hit & 0.3800 & 0.7900 & 0.6500 & 0.5500 & 0.3000 & \textbf{0.8700} \\ \midrule
\multirow{8}{*}{E-commerce} & Comprehensiveness & 23.00\% & 32.00\% & 18.50\% & 17.50\% & 15.25\% & \textbf{70.75\%} \\
 & Diversity & 17.75\% & 29.00\% & 14.00\% & 19.75\% & 11.75\% & \textbf{59.50\%} \\
 & Empowerment & 22.50\% & 30.75\% & 17.25\% & 13.25\% & 16.50\% & \textbf{60.00\%} \\
 & Directness & 28.25\% & 21.50\% & 31.25\% & 38.00\% & 15.25\% & \textbf{43.00\%} \\
 & Overall & 15.75\% & 22.50\% & 14.50\% & 12.00\% & 8.75\% & \textbf{48.75\%} \\ \cmidrule(l){2-8} 
 & F1-score & 0.2159 & 0.5694 & 0.2627 & 0.2787 & 0.1877 & \textbf{0.5887} \\
 & Hit & 0.3400 & \textbf{0.9100} & 0.4600 & 0.5300 & 0.3300 & 0.8700 \\ \bottomrule
\end{tabular}%
}
\end{table}

\stitle{Decomposition over datasets.} In~\autoref{tab:quality_dataset}, we report the detailed win rates of the methods with Claude-3.5-sonnet, averaged across all question patterns for each knowledge graph. The results show that \name{} consistently outperforms the baselines on all three knowledge graphs, achieving the highest \textit{Win Rate}, \textit{F1-score}, and \textit{Hit} in 18 out of 21 cases. Notably, the win rates of \name{} on the Academia graph are generally higher, as this graph features more clearly defined entity types (e.g., \textit{author}, \textit{paper}, and \textit{venue}) and well-structured relations (e.g., \textit{reference} and \textit{cited by}), making it easier for \name{} to generate appropriate Cypher queries and retrieve the desired information.

\subsection{Response Latency and Token Usage Comparison}

\begin{table}[ht]
\caption{Response latency and token usage of the methods on each question pattern averaged across datasets. The results are in the form \textit{time/tokens.} \textbf{Bold faces} denote the lowest response latency and token usage.}
\label{tab:latency_token_pattern}
\resizebox{\textwidth}{!}{%
\begin{tabular}{@{}crrrrrr@{}}
\toprule
\textbf{Question Pattern} & \textbf{MS\_GraphRAG} & \textbf{RoG} & \textbf{Fast-graphrag} & \textbf{Graph-CoT} & \textbf{Cypher} & \textbf{PolyG} \\ \midrule
$\langle s,*,* \rangle$ & \textbf{9.78}/6,712 & 14.38/\textbf{1,142} & 39.17/53,263 & 46.30/39,381 & 15.89/4,975 & 11.19/8,210 \\
$\langle s,p,* \rangle $ & \textbf{4.85}/5,756 & 13.21/\textbf{1,556} & 36.44/44,926 & 44.49/55,782 & 9.86/2,929 & 20.26/4,004 \\
$\langle s,*,o \rangle $ & \textbf{8.01}/14,557 & 16.84/\textbf{1,019} & 35.99/48,928 & 51.93/64,534 & 15.16/2,700 & 15.18/3,154 \\
$\langle s,p,o \rangle $ & 27.51/38,424 & 17.90/\textbf{1,819} & 34.85/44,900 & 69.93/254,260 & \textbf{12.96}/5,045 & 19.82/4,069 \\
Nested & \textbf{9.93}/21,875 & 19.00/\textbf{1,550} & 35.97/48,175 & 70.85/109,016 & 13.82/5,720 & 57.79/15,314 \\ \bottomrule
\end{tabular}%
}
\end{table}

\begin{table}[ht]
\caption{Response latency and token usage of the methods on each question pattern and each knowledge graph. The results are in the form \textit{time/tokens.} \textbf{Bold faces} denote the lowest response latency and token usage.}
\label{tab:latency_token_all}
\resizebox{\textwidth}{!}{%
\begin{tabular}{@{}ccrrrrrr@{}}
\toprule
\textbf{Dataset} & \textbf{Question Pattern} & \textbf{MS\_GraphRAG} & \textbf{RoG} & \textbf{Fast-graphrag} & \textbf{Graph-CoT} & \textbf{Cypher} & \textbf{PolyG} \\ \midrule
\multirow{5}{*}{Academia} & $\langle s,*,* \rangle$ & \textbf{11.22}/8,664 & 12.2/\textbf{1,028} & 32.23/73,089 & 50.7/41,054 & 13.88/3,111 & 12.56/10,163 \\
 & $\langle s,p,* \rangle$ & \textbf{4.87}/8,043 & 13.06/\textbf{1,328} & 28.08/64,863 & 36.21/40,710 & 8.89/3,387 & 27.9/4,977 \\
 & $\langle s,*,o \rangle$ & \textbf{7.10}/13,613 & 16.98/\textbf{935} & 29.61/72,270 & 50.61/55,243 & 16.97/2,317 & 30.61/3,464 \\
 & $\langle s,p,o \rangle$ & 33.48/23,391 & 22.03/\textbf{1,759} & 26.07/55,755 & 58.3/179,926 & \textbf{10.31}/5,758 & 28.58/5,771 \\
 & Nested & \textbf{8.21}/13,903 & 19.27/\textbf{1,825} & 30.44/73,834 & 60.42/74,518 & 15.41/11,684 & 56.47/19,911 \\ \midrule
\multirow{5}{*}{Literature} & $\langle s,*,* \rangle$ & \textbf{9.47}/4,457 & 15.11/\textbf{1,244} & 18.35/25,479 & 48.46/43,434 & 13.03/2,502 & 10.68/5,951 \\
 & $\langle s,p,* \rangle$ & \textbf{4.26}/1,931 & 13.07/\textbf{1,781} & 17.69/28,755 & 50.9/77,353 & 9.95/3,078 & 17.2/3,791 \\
 & $\langle s,*,o \rangle$ & 8.9/12,261 & 14.73/\textbf{1,099} & 17.05/28,905 & 53.35/83,293 & 11.78/1,357 & \textbf{6.55}/2,781 \\
 & $\langle s,p,o \rangle$ & 31.54/53,826 & 13.72/\textbf{1,547} & 18.11/38,993 & 87.83/413,733 & \textbf{8.78}/1,602 & 14.65/3,276 \\
 & Nested & \textbf{8.20}/17,395 & 19.93/\textbf{1,396} & 17.65/30,094 & 71.9/104,617 & 11.14/1,575 & 69.82/14,714 \\ \midrule
\multirow{5}{*}{E-commerce} & $\langle s,*,* \rangle$ & \textbf{8.66}/7,016 & 15.83/\textbf{1,153} & 66.93/61,220 & 39.73/33,655 & 20.76/9,313 & 10.32/8,516 \\
 & $\langle s,p,* \rangle$ & \textbf{5.42}/7,294 & 13.51/\textbf{1,558} & 63.54/41,161 & 46.35/49,283 & 10.74/2,322 & 15.67/3,244 \\
 & $\langle s,*,o \rangle$ & \textbf{8.04}/17,797 & 18.81/\textbf{1,024} & 61.32/45,610 & 51.84/55,067 & 16.74/4,426 & 8.39/3,218 \\
 & $\langle s,p,o \rangle$ & 17.5/38,054 & 17.95/\textbf{2,150} & 60.36/39,952 & 63.67/169,121 & 19.8/7,775 & \textbf{16.24}/3,160 \\
 & Nested & \textbf{13.38}/34,328 & 17.81/\textbf{1,429} & 59.82/40,598 & 80.24/147,912 & 14.91/3,902 & 47.08/11,316 \\ \bottomrule
\end{tabular}%
}
\end{table}

\stitle{Decomposition over question patterns.} \autoref{tab:latency_token_pattern} presents the detailed response latency and token usage of \name{} and the baseline methods across each question pattern using Claude-3.5-sonnet, averaged over the three knowledge graphs. The results show that, in most cases, MS\_GraphRAG achieves the lowest response latency, while RoG consumes the fewest tokens. However, this low execution complexity comes at the cost of generation quality. For instance, although MS\_GraphRAG is fast, it only performs best in 3 out of 31 criteria for generation quality, as shown in~\autoref{tab:quality_decompose}. Similarly, RoG uses the least tokens but only wins in 2 out of 31 criteria. Since response quality is the primary factor that determines the usefulness of an answer, their low latency or token usage does not indicate stronger capability for general GraphRAG workloads.

Meanwhile, although \name{} does not always offer the best execution efficiency or lowest resource consumption, its response latency and token usage are comparable to the best-performing baselines in most cases (e.g., 11.19s vs. 9.78s in latency and 3,154 vs. 1,019 tokens). Moreover, across all question patterns, \name{} achieves up to 4× speedup in response latency (e.g., for $\langle s,*,* \rangle$) and reduces token usage by more than 90\% (e.g., for $\langle s,p,o \rangle$) compared to the worst-performing baselines.

\stitle{Decomposition over datasets.} \autoref{tab:latency_token_all} further decomposes the results from~\autoref{tab:latency_token_pattern} by breaking them down by each knowledge graph, showing the response latency and token usage of each method across different question patterns and datasets. In addition to the similar observations made in~\autoref{tab:latency_token_pattern}, we find that \name{} achieves the lowest response latency for $\langle s,*,o \rangle$ questions on the Literature graph and for $\langle s,p,o \rangle$ questions on the E-commerce graph. Furthermore, response latency and token usage are generally higher on the Academia graph, which can be attributed to its dense graph topology, especially in the \textit{author-paper} and \textit{paper-paper} relations.


\section{Related Work}\label{sec:existing_work_appendix}

\begin{figure*}[!ht]
	\centering
\includegraphics[width=\textwidth]{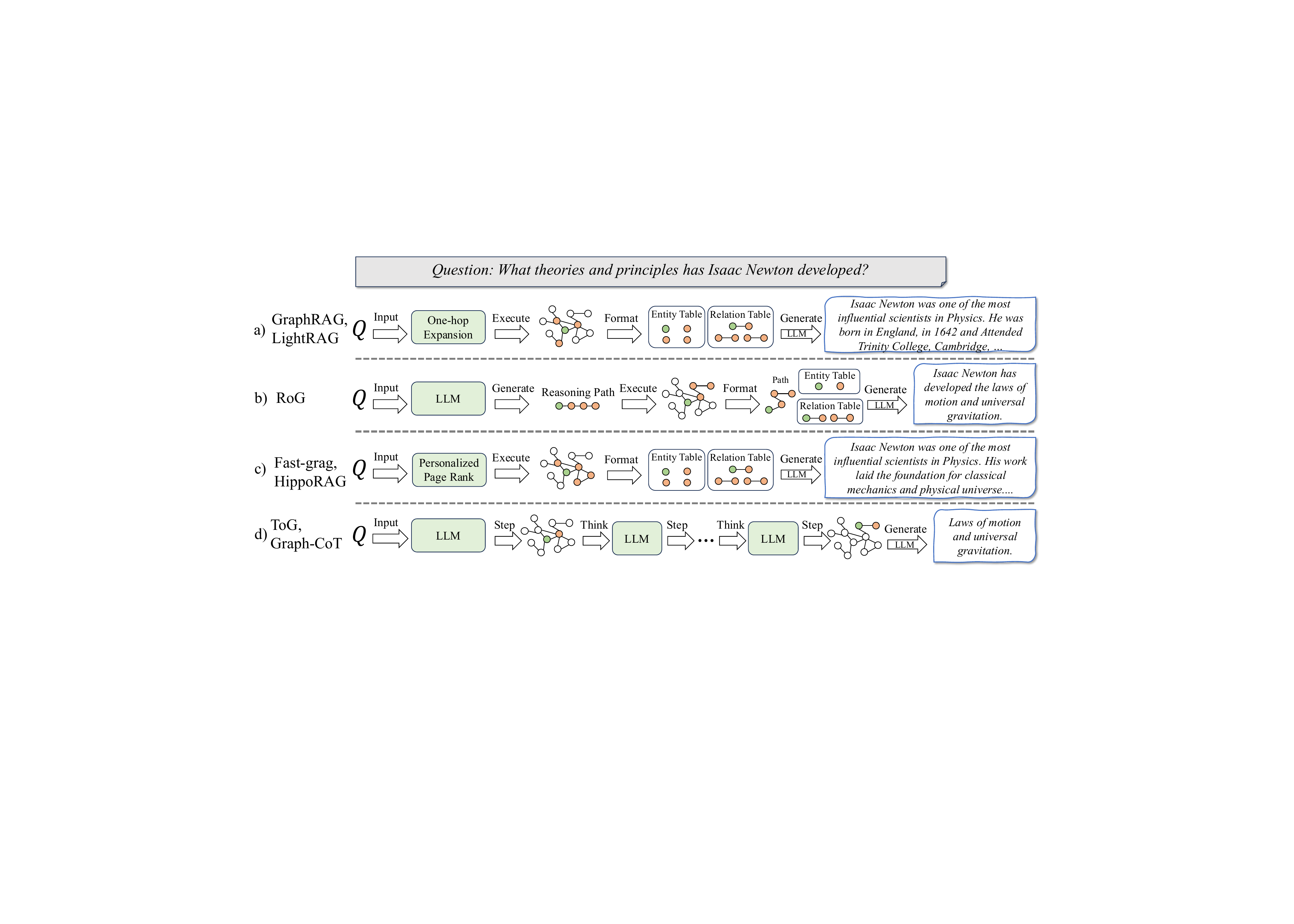}
	\caption{Workflow demonstration and comparison of existing GraphRAG methods.}
	\label{fig:existing_work}
\end{figure*}

Many GraphRAG approaches are proposed to tackle KGQA tasks or answer general user questions about the knowledge graph, and they differ in the adopted graph traversal methods. In the following, we introduce representative GraphRAG methods and \autoref{fig:existing_work} illustrates the workflow of them.

As shown in \autoref{fig:existing_work}a, GraphRAG~\cite{ms_graphrag} and LightRAG~\cite{lightrag} use one-hop neighbor expansion to retrieve neighboring entities and relations of the anchor entities (entities that appear in the question), and return two tables as the retrieved context to the LLM as external grounding knowledge. While one-hop neighbor expansion is efficient in execution, this method fails to capture long-distance knowledge and thus cannot well answer questions that require reasoning paths beyond one-hop.

RoG~\cite{rog}, as depicted in \autoref{fig:existing_work}b, prompts the LLM to generate faithful reasoning paths that can be followed on the knowledge graph and returns the retrieved paths for generation. Faithful reasoning paths directly point from the anchor entity to the answer entities without involving useless noise in the context. However, it is only applicable when the question itself explicitly reveals concrete relation constraints. If the question is abstract in its predicate chain, e.g., asking about general information of an entity or relationships with another entity, no faithful reasoning paths could be given.

The workflow of Fast-graphrag~\cite{fast_graphrag} and HippoRAG~\cite{hipporag} is shown in \autoref{fig:existing_work}c. It retrieves entities and relations by running the well-studied random walk method, Personalized PageRank, and returns those with a relevance score greater than a threshold. Due to the diversity and complexity of graph questions, this fuzzing relevance matching that purely depends on the structure of knowledge graph might not accurately find what the question exactly requires.

The workflow of ToG~\cite{tog} and Graph-CoT~\cite{graphcot} is displayed in \autoref{fig:existing_work}d. They hand over the decision on every step of graph traversal to the LLM, prompting it with the current neighbors and all previous contexts to determine which neighbors to explore next. Traversal stops when the LLM thinks it reaches the answer, and then the answer entities are returned as results. This chain-of-thought manner is adaptable to various cases by leveraging the reasoning adaptability in LLMs. However, they end up at high cost in both response latency and token usage, which is expensive to use in real applications.
\section{Question Pattern Identification in Existing KGQA Benchmarks}\label{sec:kbqa_benchmark_appendix}

\subsection{Introduction of Existing KGQA Benchmarks}

As listed in~\autoref{tab:benchmark_comparison}, there are seven major KGQA benchmarks developed over the past decade. Specifically, SimpleQ~\cite{simpleq} and WebQSP~\cite{webqsp} are two early KGQA benchmarks. SimpleQ constructs queries from Freebase facts by using the subjects and relationships as inputs and treating the objects as answers, which means it only contains questions of the $\langle s,p,* \rangle$ type in our taxonomy (discussed below). WebQSP is derived by selecting only those questions from WebQuestions~\cite{webquestion} that can be translated into simple SPARQL queries; the original questions in WebQuestions were generated using the Google Suggest API~\cite{googlesuggestapi}. To verify the question patterns in WebQSP, we prompt the LLM to analyze the queries and validate that most of the questions in WebQSP also belong to the $\langle s,p,* \rangle$ pattern. CWQ~\cite{cwq}, GraphQ~\cite{graphq}, and GrailQA~\cite{grailqa} are more recent benchmarks that feature higher query complexity. CWQ builds upon WebQSP by applying additional fact constraints (e.g., conjunctions and compositions) to the original SPARQL queries. GraphQ and GrailQA employ a graph-expansion algorithm to induce a subgraph from the question entity and then generate a question based on each subgraph. Although these datasets contain questions with multiple fact constraints and include features like counting, superlatives (e.g., argmax, argmin), and comparatives, the questions still ask about a concrete entity and can largely be categorized as combinations of $\langle s,p,* \rangle$ patterns.

\subsection{Question Patterns in Existing KGQA Benchmarks}

\begin{table}[ht]
\centering
\caption{Number of questions of each question pattern in WebQSP, CWQ and WIKIQA.}
\label{tab:benchmark_question_pattern}
\resizebox{0.9\textwidth}{!}{%
\begin{tabular}{@{}cccccccccc@{}}
\toprule
\multirow{3}{*}{\textbf{Benchmark}} & \multicolumn{4}{c}{\textbf{basic Question Pattern}} & \multicolumn{4}{c}{\textbf{Nested Question Pattern}} \\ \cmidrule(lr){2-9}
 & \multirow{2}{*}{$\langle s,*,* \rangle$} & \multirow{2}{*}{$\langle s,p,* \rangle$} & \multirow{2}{*}{$\langle s,*,o \rangle$} & \multirow{2}{*}{$\langle s,p,o \rangle$} & $\langle s,*,* \rangle$ & $\langle s,p,* \rangle$ & $\langle s,*,o \rangle$ & $\langle s,p,o \rangle$ &  \\
 &  &  &  &  & $\langle s,p,* \rangle$ & $\langle s,p,* \rangle$ & $\langle s,p,* \rangle$ & $\langle s,p,* \rangle$ &  \\ \midrule
WebQSP~\cite{webqsp} (2016) & 179 & 4,558 &  &  &  &  &  & \\
CWQ~\cite{cwq} (2018) &  & 19,631 &  &  &  & 15,058 &  & \\ 
WIKIQA~\cite{wikiqa} (2015) & 761 & 1,753 & 276 & & 24 & 224 & & \\ \bottomrule
\end{tabular}%
}
\end{table}

For most benchmarks, the question pattern can be inferred from their question generation procedures—for example, from the templates used in RGBench~\cite{graphcot} and CypherBench~\cite{cypherbench}, or the subgraph construction methods in GraphQ~\cite{graphq} and GrailQA~\cite{grailqa}. However, for WebQSP~\cite{webqsp} and CWQ~\cite{cwq}, we need to examine the questions directly to determine their corresponding patterns, as these questions are generated using the Google Suggest API. To this end, we classify each question in WebQSP and CWQ using an LLM (Claude-3.5-sonnet). The prompt used to guide the LLM in identifying the question pattern is largely the same as the one used in \name{} for question categorization, as shown in~\autoref{fig:qc_prompt}. This prompt provides clear definitions of each question pattern, includes concrete few-shot examples, and allows for exceptions where questions do not fit cleanly into one of the four basic patterns and should instead be labeled as nested questions. Additionally, we ask the LLM to provide a question decomposition for each nested question. In the following, we discuss the question categorization results of WebQSP and CWQ, respectively.

\autoref{tab:benchmark_question_pattern} reports the results on WebQSP and CWQ. We observe that the majority of questions in the WebQSP benchmark fall into the $\langle s,p,*\rangle$ pattern, with only 179 out of 4,737 questions belonging to the $\langle s,*,* \rangle$ category. An example of a $\langle s,*,* \rangle$ question is “What is Mount St. Helens?”. Although such questions typically lack definitive ground-truth answers, WebQSP annotates responses like “Stratovolcano” and “Volcano”, which reduces the evaluation to checking for the presence of certain keywords rather than assessing the overall quality or helpfulness of the information. There are no $\langle s,*,o \rangle$ or $\langle s,p,o \rangle$ questions in WebQSP, as it is difficult to define a unique set of ground-truth answers for these types. In contrast, CWQ~\cite{cwq}, which is sampled and augmented from WebQSP, introduces additional $\langle s,p,* \rangle + \langle s,p,* \rangle$ questions by incorporating extra fact constraints using conjunctions, superlatives, comparatives, and compositions. This results in 15,058 nested $\langle s,p,* \rangle + \langle s,p,* \rangle$ questions out of the total 34,689 questions.

To study the diversity and possible graph question patterns that may arise in real-world scenarios, we analyze WIKIQA~\cite{wikiqa}, a text-understanding benchmark whose questions are extracted from real Bing query logs. We apply the same classification method used for WebQSP and CWQ to determine the pattern of each question in WIKIQA. As reported in~\autoref{tab:benchmark_question_pattern}, among the 3,047 questions, 761 fall under the $\langle s,*,* \rangle$ pattern, 1,753 under $\langle s,p,* \rangle$, and 276 under $\langle s,*,o \rangle$. Among the nested questions, 24 follow the $\langle s,*,* \rangle + \langle s,p,* \rangle$ pattern and 224 follow $\langle s,p,* \rangle + \langle s,p,* \rangle$. No $\langle s,p,o \rangle$ questions are found in WIKIQA, likely because the dataset only includes question-like queries that begin with WH-words (e.g., “what” or “how”) and end with a question mark. In contrast, $\langle s,p,o \rangle$ questions are more likely to start with expressions such as “Is there,” “Do,” or “Have.” Furthermore, since WIKIQA extracts only a subset of Bing query log data using basic word filtering, its question pattern distribution may not fully represent real-world scenarios. Nonetheless, the presence of multiple question patterns in WIKIQA supports the validity of our proposed taxonomy.

\begin{figure*}[!h]
	\centering
\includegraphics[width=\textwidth]{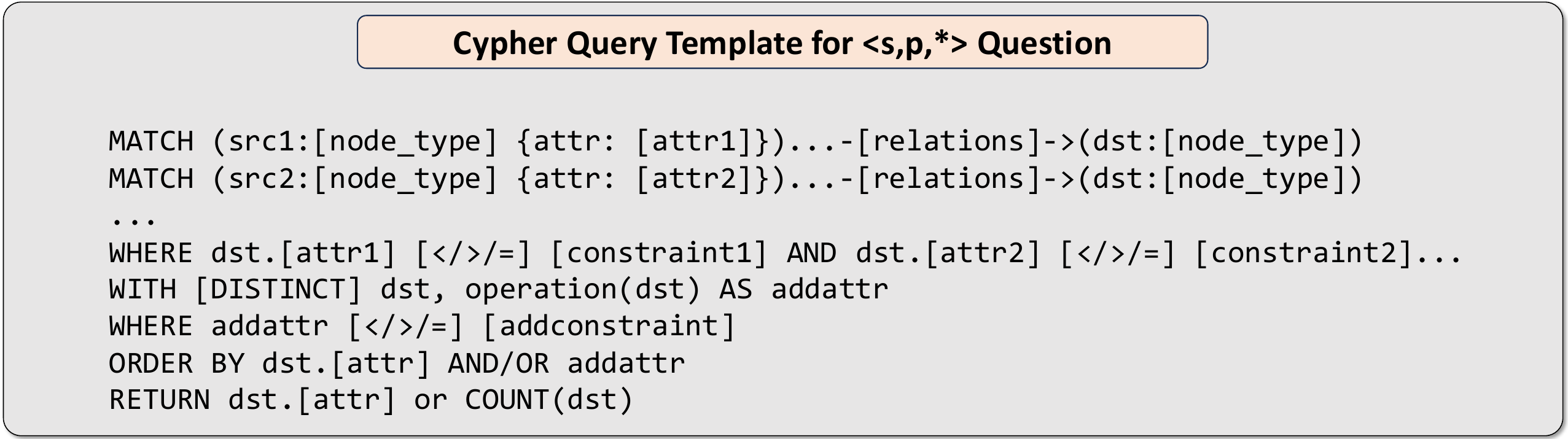}
	\caption{Cypher template for $\langle s,p,* \rangle$ questions.}
	\label{fig:cyper_template}
\end{figure*}

As we have established, most existing KGQA benchmarks primarily contain questions following the $\langle s,p,* \rangle$ pattern and its nested extension $\langle s,p,* \rangle + \langle s,p,* \rangle$. We now illustrate how representative questions from these benchmarks exhibit strong similarities and limit the diversity of question coverage. In particular, we observe that they share similar Cypher query structures for retrieving the desired answers, which can be abstracted into a general template, as shown in~\autoref{fig:cyper_template}. Empirically, all $\langle s,p,* \rangle$ questions can be instantiated using this Cypher template, with the specific graph schema guiding how the relations in the graph correspond to the fact constraints expressed in the user question. Below, we present several examples to demonstrate how this template can be instantiated for various $\langle s,p,* \rangle$ questions.

\circled{1} \textit{"Which city held the summer Olympics twice?"}

Cypher query:

\begin{lstlisting}[style=cypher, language=]
MATCH (src:[Event] {name: [Summer Olympics]})-[held in]->(dst:[City])
WITH DISTINCT dst, COUNT(dst) AS count
WHERE count == 2
RETURN dst.name
\end{lstlisting}

\circled{2} \textit{"What year did the Florida Marlins win their 2nd world series title?"}

Cypher query:

\begin{lstlisting}[style=cypher, language=]
MATCH (src:[Title] {name: [World Series Chaimpion]})-[rel:won by]->(dst:[Team])
WHERE dst.name == Florida Marlins
WITH dst, src
ORDER BY src.date
RETURN dst.name
LIMIT 1
\end{lstlisting}

\circled{3} \textit{"Which states does the river that flows under the DeSoto Bridge pass through?"}

Cypher query:

\begin{lstlisting}[style=cypher, language=]
MATCH (src:[Bridge] {name: [DeSoto Bridge]})-[on top of]->(n:[River])-[pass through]->(n:[City])-[belong to]->(dst:[States])
WITH DISTINCT dst
RETURN dst.name
\end{lstlisting}

\circled{4} \textit{"How many languages are used where \"Lupang Hinirang\" is the national anthem?"}

Cypher query:

\begin{lstlisting}[style=cypher, language=]
MATCH (src:[Song] {name: [Lupang Hinirang]})-[national anthem]->(n:[Country])-[people]->(n:[People])-[speak]->(dst:[Language])
WITH DISTINCT dst
RETURN COUNT(dst)
\end{lstlisting}

\section{Dataset Used in \benchmarkname}\label{sec:dataset_appendix}

\begin{table}[ht]
\centering
\caption{Statistics of the knowledge graphs used for evaluation}
\label{tab:datasets}
\begin{tabular}{@{}cccc@{}}
\toprule
\textbf{Attributes}     & \textbf{Academia} & \textbf{Literature} & \textbf{E-commerce} \\ \midrule
\# of Entities       & 2.7M             & 3.7M               & 8.1M            \\
\# of Relations      & 67M              & 28M                & 193M            \\
\# Templates      & 24              & 28                & 21  \\
\# of Questions      & 400              & 400                & 400  \\ \bottomrule
\end{tabular}%
\end{table}

The statistics of the datasets used in \benchmarkname{} are presented in~\autoref{tab:datasets}. The three knowledge graphs are sourced from RGBench~\cite{graphcot}. Specifically, in the academic domain, there are three types of entities: \textit{papers}, \textit{authors}, and \textit{venues}. These are naturally interconnected by citation-related relationships such as ``\textit{written-by}'' and ``\textit{publish-in}''. The selected academic graph is from the Physics domain of DBLP~\cite{academia}. 

In the literature domain, the knowledge graph contains four types of entities—\textit{books}, \textit{authors}, \textit{publishers}, and \textit{series}. Its structure captures the publication relationships among these entities, extracted from the Goodreads dataset~\cite{literature}, which provides a rich collection of books along with detailed metadata. The relations include ``\textit{written-by}'', ``\textit{publish-in}'', ``\textit{book-series}'', and others.

For the e-commerce domain, we adopt the graph from the Amazon product dataset~\cite{e-commerce}, which offers metadata for items across a wide range of product categories. In this domain, the nodes include \textit{items} and \textit{brands}, and the interconnecting relations include ``\textit{also-viewed}``, ``\textit{also-bought}'', ``\textit{buy-after-viewing}'', ``\textit{bought-together}'' and ``\textit{item-brand}''.

In the following, we list the concrete graph schema (i.e., entity and relation types) in the above three knowledge graphs.

\subsection{Academia Domain}

\textit{Node properties:}
\squishlist
\item type: author, properties: ["id", "name", "node\_type"]
\item type: paper, properties: ["id", "name", "label", "year", "node\_type", "abstract"]
\item type: venue, properties: ["id", "name", "node\_type"]
\squishend

\textit{Edge properties:}

Author nodes are linked to their paper nodes by authorship.
\squishlist
\item author $\rightarrow$ "paper" $\rightarrow$ paper
\squishend

Paper nodes are linked to their author nodes, venue nodes, other paper nodes.
\squishlist
\item paper $\rightarrow$ "author" $\rightarrow$ author
\item paper $\rightarrow$ "reference" $\rightarrow$ paper
\item paper $\rightarrow$ "venue" $\rightarrow$ venue
\item paper $\rightarrow$ "cited\_by" $\rightarrow$ paper
\squishend

Venue nodes are linked to their included paper nodes.
\squishlist
\item venue $\rightarrow$ "paper" $\rightarrow$ paper
\squishend

\subsection{Literature Domain}

\textit{Node properties:}
\squishlist
\item type: book, properties: ["id", "name", "node\_type", "description", "publication\_year", "genres"]
\item type: author, properties: ["id", "name", "node\_type"]
\item type: publisher, properties: ["id", "name", "node\_type"]
\item type: series, properties: ["id", "name", "node\_type", "description"]
\squishend

\textit{Edge properties:}

Book nodes are linked to book nodes, author nodes, publisher nodes and series nodes.
\squishlist
\item book $\rightarrow$ "author" $\rightarrow$ author
\item book $\rightarrow$ "publisher" $\rightarrow$ publisher
\item book $\rightarrow$ "series" $\rightarrow$ series
\item book $\rightarrow$ "similar\_books" $\rightarrow$ book
\squishend

Author nodes are linked to their neighboring book nodes.
\squishlist
\item author $\rightarrow$ "book" $\rightarrow$ book
\squishend

Publisher nodes are linked to their neighboring book nodes.
\squishlist
\item publisher $\rightarrow$ "book" $\rightarrow$ book
\squishend

Series nodes are linked to their neighboring book nodes.
1. series $\rightarrow$ "book" $\rightarrow$ book

\subsection{E-commerce Domain}

\textit{Node properties:}
\squishlist
\item type: item, properties: ["id", "name", "node\_type"]
\item type: brand, properties: ["id", "name", "node\_type"]
\squishend

Edge properties:
Item nodes are linked to neighboring item nodes and brand nodes.
\squishlist
\item item $\rightarrow$ "also\_viewed\_item" $\rightarrow$ item
\item item $\rightarrow$ "buy\_after\_viewing\_item" $\rightarrow$ item
\item item $\rightarrow$ "also\_bought\_item" $\rightarrow$ item
\item item $\rightarrow$ "bought\_together\_item" $\rightarrow$ item
\item item $\rightarrow$ "brand" $\rightarrow$ brand
\squishend

Brand nodes are linked to their neighboring item nodes. Specific relations are:
\squishlist
\item brand $\rightarrow$ "item" $\rightarrow$ item
\squishend
\section{Question Templates for Question Generation}\label{sec:question_template_appendix}

In the following, we list the question templates used in question generation for each knowledge graph.

\subsection{Academia Domain}

\stitle{$\langle s,*,*\rangle $ Question Template.}

\squishlist
\item \textit{Give me a broad introduction about "[author name]".}
\item \textit{Tell me some information about "[paper name]".}
\item \textit{Give me a comprehensive description about "[venue name]".}
\squishend

\stitle{$\langle s,p,*\rangle $ Question Template.}

\squishlist
\item \textit{What paper have the author "[author name]" published?}
\item \textit{What are the academic collaborators of "[author name]"?}
\item \textit{What venues have the author "[author name]" published in?}
\item \textit{Who are the authors of the paper "[paper name]"?}
\item \textit{Where is the paper "[paper name]" published?}
\item \textit{Who are the academic collaborators of the author who writes the paper "[paper name]"?}
\item \textit{What venues have the author of the paper "[paper name]" published in?}
\item \textit{What venues have the academic collaborators of the author who writes the paper "[paper name]" published in?}
\squishend

\stitle{$\langle s,*,o\rangle $ Question Template.}

\squishlist
\item \textit{How does "[author name]" and "[author name]" influence each other?}
\item \textit{How are "[paper name]" and "[paper name]" related to each other?}
\item \textit{What is the connection between "[author name]" and "[paper name]"?}
\item \textit{What is the relationship between "[venue name]" and "[paper name]"?}
\item \textit{how are "[venue name]" and "[author name]" connected?}
\squishend

\stitle{$\langle s,p,o\rangle $ Question Template.}

\squishlist
\item \textit{Have the author "[author name]" cited or been cited by the work of the author "[author name]" and what are those works?}
\item \textit{Have authors "[author name]" and "[author name]" both collaborated with some other authors and who are they?}
\item \textit{Have the authors "[author name]" and "[author name]" ever published papers in the same venues? If so, tell me some examples.}
\item \textit{Do the venues "[venue name]" and "[venue name]" have the same authors publishing work in both of them and who are they?}
\squishend

\stitle{Nested Question Template.}

\squishlist
\item \textit{Tell me about the academic contributions of the academic collaborators of the scholar "[author name]".}
\item \textit{Who are the academic collaborators of the author who writes both the paper "[paper name]" and paper "[paper name]"?}
\item \textit{What is the relationship between the scholar '{}' and the authors of the paper "[paper name]"?}
\item \textit{Have the scholars "[author name]" and "[author name]" both published work at the venue having the paper "[paper name]"?}
\squishend

\subsection{Literature Domain}

\stitle{$\langle s,*,*\rangle $ Question Template.}

\squishlist
\item \textit{Tell me some information about "[author name]".}
\item \textit{Give me a broad introduction about "[book name]".}
\item \textit{Briefly describe "[publisher name]".}
\item \textit{Give me a comprehensive description about "[series name]".}
\squishend

\stitle{$\langle s,p,*\rangle $ Question Template.}

\squishlist
\item \textit{Who are the authors of the book "[book name]"?}
\item \textit{What series have the author of the book "[book name]" published?}
\item \textit{What books has the author "[author name]" published?}
\item \textit{What books have the collaborators of the author "[author name]" published?}
\item \textit{What are the series published by the publishers that have published books of the author "[author name]"?}
\item \textit{What are the authors of the books published by the publisher "[publisher name]"?}
\item \textit{Where does the books of the series "[series name]" published in?}
\item \textit{What are the authors of the books that are published by the publishers that have published books of the series "[series name]"?}
\squishend

\stitle{$\langle s,*,o\rangle $ Question Template.}

\squishlist
\item \textit{How do "[author name]" and "[author name]" influence each other?}
\item \textit{How is "[author name]" related to "[book name]"?}
\item \textit{How are "[book name]" and "[book name]" connected?}
\item \textit{What is the connection between "[book name]" and "[publisher name]"?}
\item \textit{What is the relationship between "[book name]" and "[series name]"?}
\item \textit{How are "[publisher name]" and "[publisher name]" related to each other?}
\item \textit{What is the relationship between "[series name]" and "[series name]"?}
\item \textit{What are "[series name]" and "[publisher name]" related?}
\squishend

\stitle{$\langle s,p,o\rangle $ Question Template.}

\squishlist
\item \textit{Have the authors "[author name]" and "[author name]" ever published books in the same publishers? If so, tell me some examples.}
\item \textit{Do the publishers "[publisher name]" and "[publisher name]" have any authors publishing books in both of them and what are the publications and authors?}
\item \textit{Do the publishers "[publisher name]" and "[publisher name]" have books that belong to the same series, and if so, what are those books?}
\item \textit{Do the series "[series name]" and "[series name]" contain books that are published by the same publisher? If so, tell me about them.}
\squishend

\stitle{Nested Question Template.}

\squishlist
\item \textit{Provide a comprehensive overview about the series whose books are similar to the publications of author "[author name]"?}
\item \textit{What books are published by the collaborators of both the author "[author name]" and "[author name]"?}
\item \textit{What is the relationship between the author "[author name]" and the author who has the book "[book name]"?}
\item \textit{Have the authors "[author name]" and "[author name]" ever published books in the same publisher that has the series "[series name]"?}
\squishend

\subsection{E-commerce Domain}

\stitle{$\langle s,*,*\rangle $ Question Template.}

\squishlist
\item \textit{Tell me about "[brand name]".}
\item \textit{Tell me about "[item name]".}
\squishend

\stitle{$\langle s,p,*\rangle $ Question Template.}

\squishlist
\item \textit{What is the brand of the item "[item name]"?}
\item \textit{What are the brands of the items that are also bought after viewing the item "[item name]"?}
\item \textit{What are the items that are also viewed when viewing items of the brand owning the item "[item name]"?}
\item \textit{What are the brands of the items that are also bought with items of the brand owning the item "[item name]"?}
\item \textit{What are the items of the brand "[brand name]"?}
\item \textit{What are the items that are bought together with items of the brand "[brand name]"?}
\item \textit{What are the brands of the items that are also bought with items of the brand "[brand name]"?}
\item \textit{What items does the brands of the items that are also viewed together with items of the brand "[brand name]" have?}
\squishend

\stitle{$\langle s,*,o\rangle $ Question Template.}

\squishlist
\item \textit{What is the relationship between "[brand name]" and "[brand name]"?}
\item \textit{What are "[item name]" and "[item name]" related to each other?}
\item \textit{What are "[brand name]" and "[item name]" connected?}
\squishend

\stitle{$\langle s,p,o\rangle $ Question Template.}

\squishlist
\item \textit{Have the items of the brands "[brand name]" and "[brand name]" ever both been also bought with some other items, and if so, what are those items?}
\item \textit{Have the items of the brands "[brand name]" and "[brand name]" ever both been bought after viewing some other items, and if so, what are those items?}
\item \textit{Have the items of the brands "[brand name]" and "[brand name]" ever been viewed together with some other items, and if so, what are those items?}
\item \textit{Have the items of the brands "[brand name]" and "[brand name]" ever been bought together with some other items, and if so, what are those items?}
\squishend

\stitle{Nested Question Template.}

\squishlist
\item \textit{Give a broad introduction about the brands that are bought together with items of the brand "[brand name]"?}
\item \textit{What are the brands that are commonly viewed with the brands "[brand name]" and also bought together with the brand "[brand name]"?}
\item \textit{What is the relationship between the brand "[brand name]" and the brand which has the item "[item name]"?}
\item \textit{Have the items of the brands "[brand name]" and "[brand name]" ever been also viewed with the items that are bought together with the item "[item name]"?}
\squishend

\section{Prompts Used in \name{}.}\label{sec:prompts_appendix}

\subsection{Question Categorization Prompt}

\begin{figure}[!ht]
	\centering
	\includegraphics[width=\textwidth]{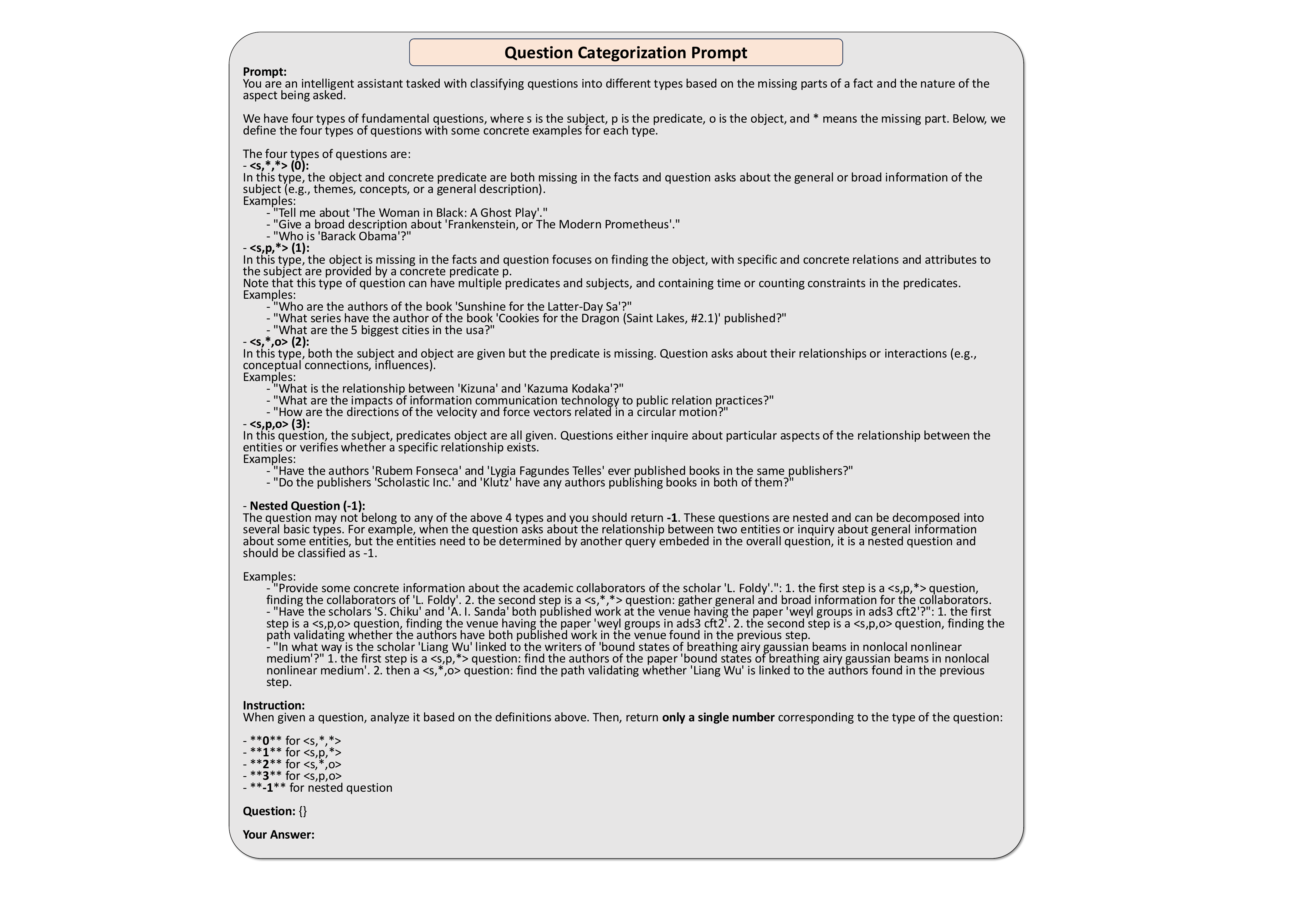}
	\caption{The prompt for the LLM to categorize questions into one of the four basic question patterns or nested questions.}
	\label{fig:qc_prompt}
\end{figure}

In \autoref{fig:qc_prompt}, we provide the prompt we use for the LLM to judge which pattern the input question belongs to, including clear definitions of each question pattern and the corresponding few-shot examples. For instance, the definition for the $\langle s,p,*\rangle $ type is "In this type, the object is missing in the facts and question focuses on finding the object, with specific and concrete relations and attributes to the subject are provided by a concrete predicate chain $p$." and we give three examples "Who are the authors of the book 'Sunshine for the Latter-Day Sa'?", "What series have the author of the book 'Cookies for the Dragon (Saint Lakes, \#2.1)' published?" and "What are the 5 biggest cities in the usa?" to the LLM to better understand how the questions in this pattern looks like.

\subsection{Question Decomposition Prompt}

\begin{figure}[!ht]
	\centering
	\includegraphics[width=\textwidth]{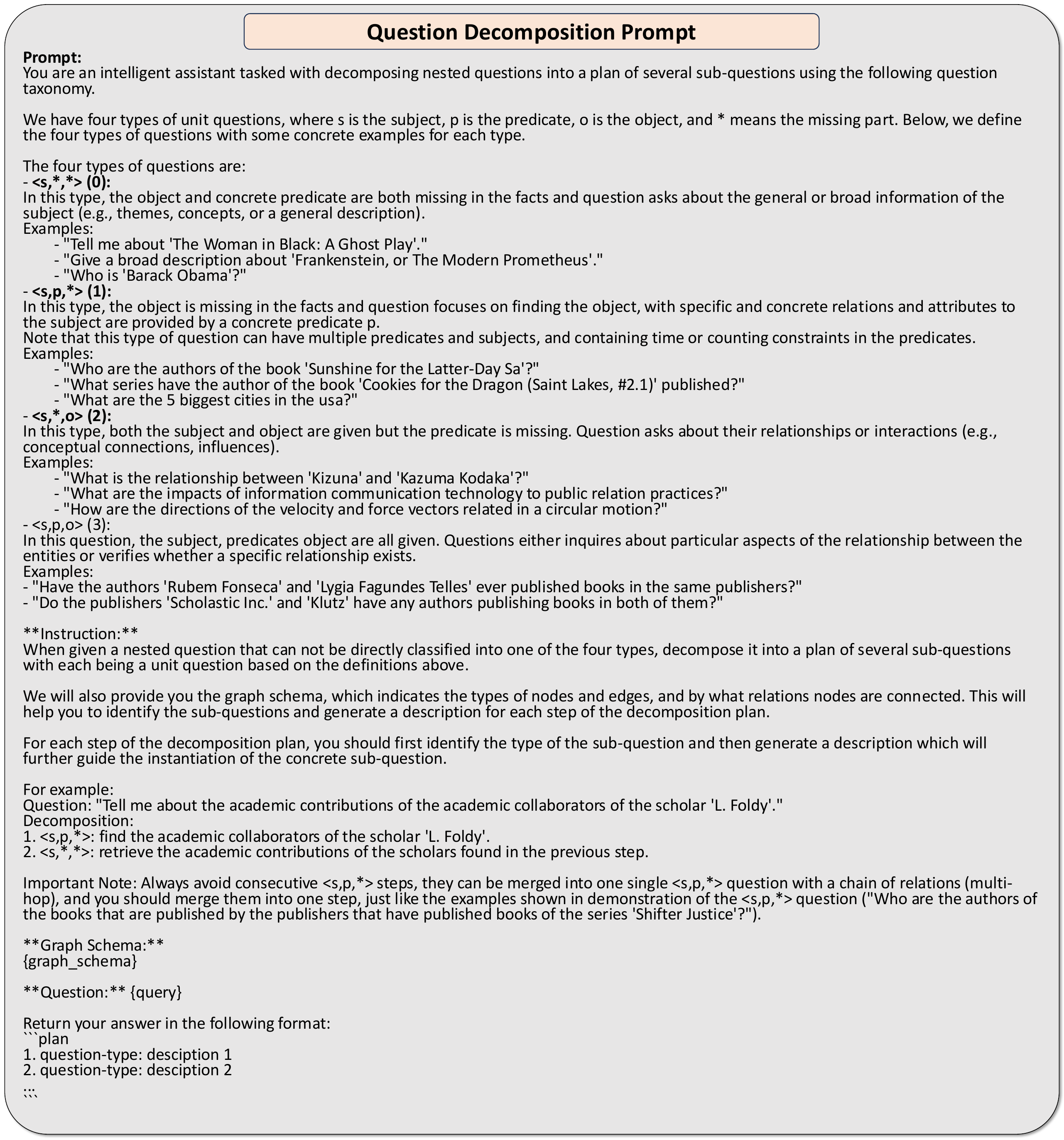}
	\caption{The prompt for the LLM to decompose nested questions into serveral of the four basic question patterns and output the detailed query plan.}
	\label{fig:qd_prompt}
\end{figure}

In \autoref{fig:qc_prompt}, we provide the prompt we use for the LLM to decompose nested questions that can not directly fall into one of the four basic question patterns. We first provide a clear definition to each basic question pattern as in~\autoref{fig:qc_prompt} with few-shot examples, and then provide the instructions for decomposition. We also provide concrete example on how to decompose nested questions in the prompt. For instance, the decomposition of the question ``Tell me about the academic contributions of the academic collaborators of the scholar 'L. Foldy'.'' is: 1) <s,p,*> question, find the academic collaborators of the scholar 'L. Foldy', 2) <s,*,*>: retrieve the academic contributions of the scholars found in the previous step. Finally, we ask the LLM to generate a detailed query plan based on given user question. We also provide the LLM with the concrete graph schema for it to generate accurate plans.

\subsection{Cypher Query Generation Prompt}

\begin{figure}[!ht]
	\centering
	\includegraphics[width=\textwidth]{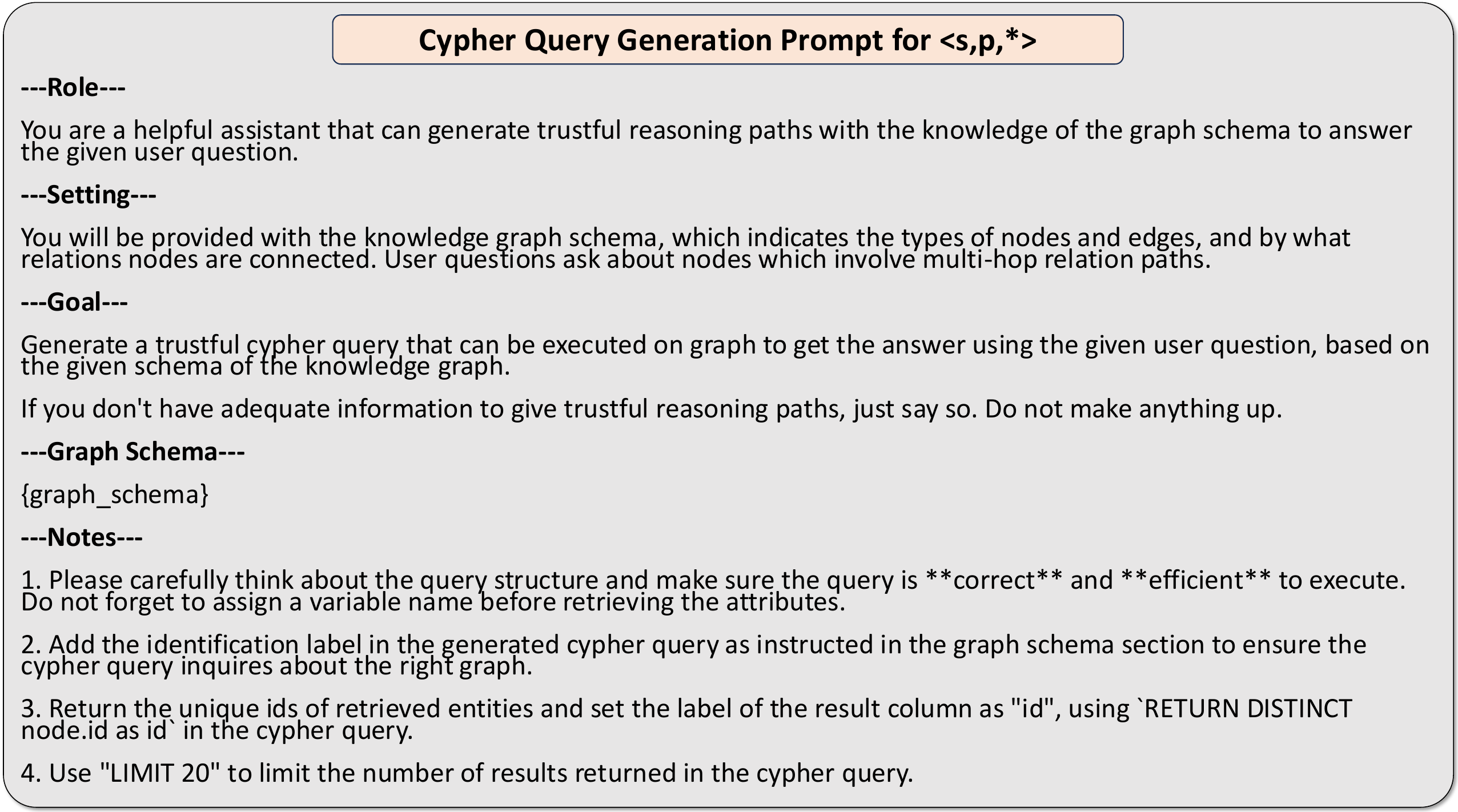}
	\caption{The prompt for the LLM to generate cypher queries for $\langle s,p,* \rangle$ questions.}
	\label{fig:spx_prompt}
\end{figure}

\begin{figure}[!ht]
	\centering
	\includegraphics[width=\textwidth]{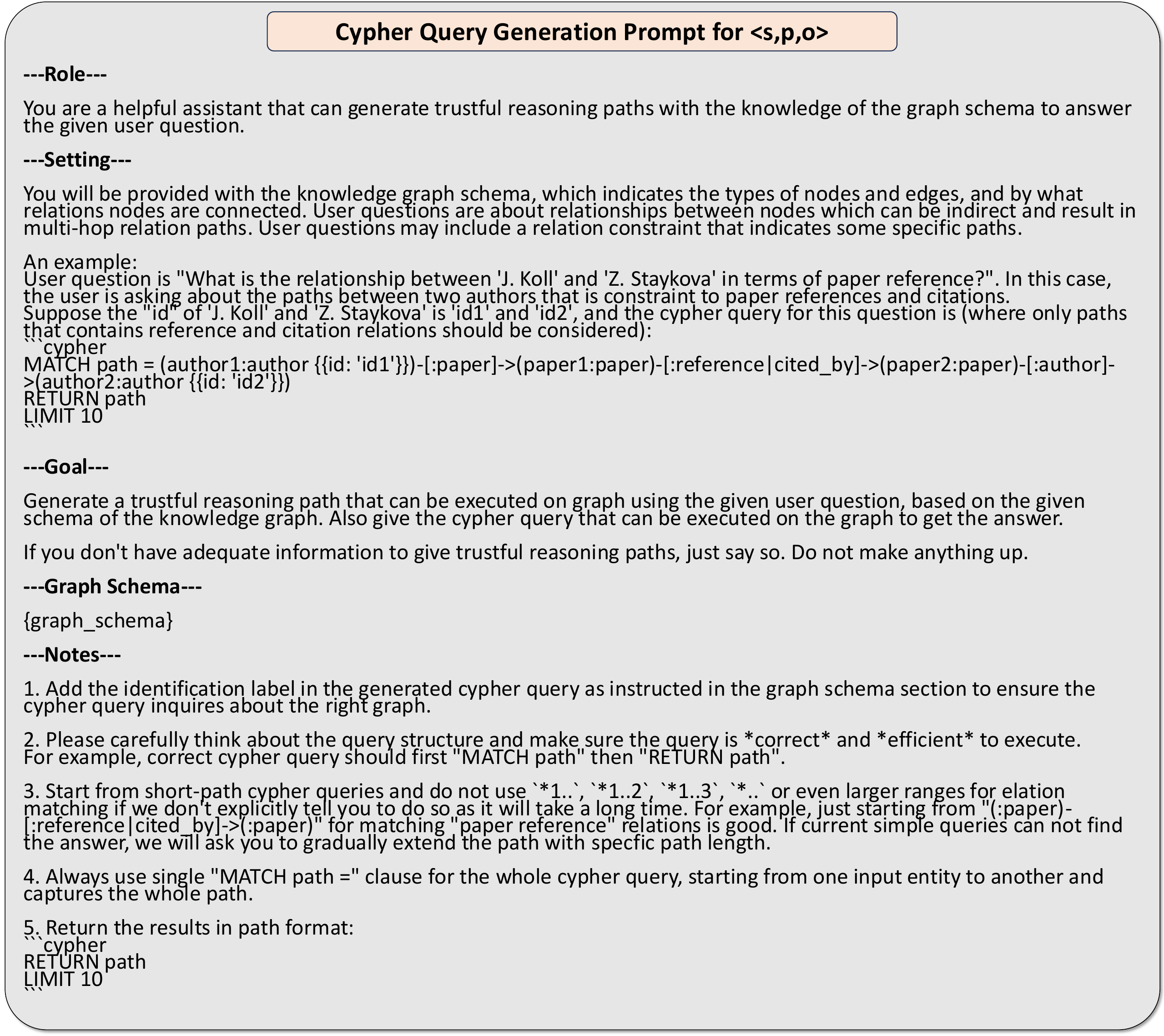}
	\caption{The prompt for the LLM to generate cypher quries for $\langle s,p,o \rangle$ questions.}
	\label{fig:spo_prompt}
\end{figure}

In~\autoref{fig:spx_prompt} and \autoref{fig:spo_prompt}, we provide the complete prompt for the LLM to generate reliable cypher queries that can directly execute in the graph database to retrieve relevant information. In particular, we first give clear specifications on the generation task and setting, with a concrete example showing the complete input and desired output. Then, we introduce the goal, which is to generate a trustful cypher query, and corresponding graph schema for reference. Lastly, we instruct the LLM with common tips to write correct and efficient cypher queries, and also specify the output format. Especially, for $\langle s,p,o \rangle$ queries, we prompt the LLM to start with short paths (i.e., lower-order relations) if there are ambiguity for matching between fact constraints in the question and the graph schema. Moreover, for $\langle s,*,* \rangle$ and $\langle s,*,o \rangle$, we directly populate the cypher query templates as mentioned in~\autoref{sec:solution} using the question entities and avoid consuming LLM tokens.

\section{Criteria Used in Evaluation}\label{sec:criteria_appendix}

The detailed explanation (prompt) of the evaluation criteria used in the experiments are as follows.

\stitle{Comprehensiveness.} How much detail does the answer provide to cover all aspects and details of the question? A comprehensive answer should be thorough and complete, without being redundant or irrelevant. For example, if the question is ’What are the benefits and drawbacks of nuclear energy?’, a comprehensive answer would provide both the positive and negative aspects of nuclear energy, such as its efficiency, environmental impact, safety, cost, etc. A comprehensive answer should not leave out any important points or provide irrelevant information. For example, an incomplete answer would only provide the benefits of nuclear energy without describing the drawbacks, or a redundant answer would repeat the same information multiple times.

\stitle{Diversity.} How varied and rich is the answer in providing different perspectives and insights on the question? A diverse answer should be multi-faceted and multi-dimensional, offering different viewpoints and angles on the question. For example, if the question is ’What are the causes and effects of climate change?’, a diverse answer would provide different causes and effects of climate change, such as greenhouse gas emissions, deforestation, natural disasters, biodiversity loss, etc. A diverse answer should also provide different sources and evidence to support the answer. For example, a single-source answer would only cite one source or evidence, or a biased answer would only provide one perspective or opinion.

\stitle{Empowerment.} How well does the answer help the reader understand and make informed judgements about the topic without being misled or making fallacious assumptions? Evaluate each answer on the quality of answer as it relates to clearly explaining and providing reasoning and sources behind the claims in the answer.

\stitle{Overall Winner.} How specifically and clearly does the answer address the question? A direct answer should provide a clear and concise answer to the question. For example, if the question is ’What is the capital of France?’, a direct answer would be ’Paris’. A direct answer should not provide any irrelevant or unnecessary information that does not answer the question. For example, an indirect answer would be ’The capital of France is located on the river Seine’.



\end{document}